\documentclass[manuscript]{acmart}
\AtBeginDocument{%
  }

\setcopyright{acmlicensed}
\copyrightyear{2018}
\acmYear{2018}
\acmDOI{XXXXXXX.XXXXXXX}
\acmConference[Conference acronym 'XX]{Make sure to enter the correct
  conference title from your rights confirmation email}{June 03--05,
  2018}{Woodstock, NY}
\acmISBN{978-1-4503-XXXX-X/2018/06}

\usepackage[table]{xcolor}
\definecolor{mikan}{RGB}{255,149,71}  
\definecolor{sakurapink}{RGB}{255,158,172}  
\definecolor{emraldgreen}{RGB}{39,193,183}  
\definecolor{myred}{RGB}{219,8,57}  
\definecolor{mylightblue}{RGB}{102,192,255} 
\definecolor{myyellow}{RGB}{248,181,0}  
\definecolor{myviolet}{RGB}{194,82,198}   
\definecolor{royalblue}{RGB}{72,94,198} 
\definecolor{milktea}{RGB}{217,219,131} 
\definecolor{senzaimidori}{RGB}{60,104,84}

\usepackage{booktabs}
\usepackage{amsmath}
\usepackage{multirow} 
\usepackage{mdframed} 
\usepackage{graphicx}
\usepackage{subcaption}
\usepackage{geometry}

\newmdenv[
  linecolor=black,
  backgroundcolor=gray!20, 
  frametitlebackgroundcolor=gray!20, 
  frametitlerule=false,
  leftmargin=10pt,
  rightmargin=10pt,
  innerleftmargin=10pt,
  innerrightmargin=10pt,
  innertopmargin=10pt,
  innerbottommargin=10pt
]{myframe}

\begin{document}

\title{Judging with Personality and Confidence: A Study on Personality-Conditioned LLM Relevance Assessment}

\author{Nuo Chen}
\email{pleviumtan@outlook.com}
\affiliation{%
  \institution{The Hong Kong Polytechnic University}
  \country{HK, China}
}

\author{Hanpei Fang}
\email{hanpeifang@ruri.waseda.jp}
\affiliation{%
  \institution{Waseda University}
  \city{Tokyo}
  \country{Japan}
}

\author{Piaohong Wang}
\affiliation{%
  \institution{City University of Hong Kong}
  \country{HK, China}
}

\author{Jiqun Liu}
\email{jiqunliu@ou.edu}
\affiliation{%
  \institution{The University of Oklahoma}
  \country{OK, USA}
}

\author{Tetsuya Sakai}
 \email{tetsuyasakai@acm.org}
\affiliation{%
  \institution{Waseda University}
    \city{Tokyo}
  \country{Japan}
}

\author{Xiao-Ming Wu}
\email{xiao-ming.wu@polyu.edu.hk}
\affiliation{%
  \institution{The Hong Kong Polytechnic University}
  \country{HK, China}
}

\renewcommand{\shortauthors}{Trovato et al.}

\begin{abstract}
Recent studies have shown that prompting can enable large language models (LLMs) to simulate specific personality traits and produce behaviors that align with those traits. However, there is limited understanding of how these simulated personalities influence critical web search decisions, specifically relevance assessment. Moreover, few studies have examined how simulated personalities impact confidence calibration, specifically the tendencies toward overconfidence or underconfidence. This gap exists even though psychological literature suggests these biases are trait-specific, often linking high extraversion to overconfidence and high neuroticism to underconfidence.

To address this gap, we conducted a comprehensive study evaluating multiple LLMs, including commercial models and open-source models, prompted to simulate Big Five personality traits. We tested these models across three test collections (TREC DL 2019, TREC DL 2020, and LLMJudge), collecting two key outputs for each query-document pair: a relevance judgment and a self-reported confidence score.

The findings show that personalities such as low agreeableness consistently align more closely with human labels than the unprompted condition. Additionally, low conscientiousness performs well in balancing the suppression of both overconfidence and underconfidence. We also observe that relevance scores and confidence distributions vary systematically across different personalities. Based on the above findings, we incorporate personality-conditioned scores and confidence as features in a random forest classifier. This approach achieves performance that surpasses the best single-personality condition on a new dataset (TREC DL 2021), even with limited training data. These findings highlight that personality-derived confidence offers a complementary predictive signal, paving the way for more reliable and human-aligned LLM evaluators.
\end{abstract}



\received{20 February 2007}
\received[revised]{12 March 2009}
\received[accepted]{5 June 2009}

\maketitle


\section{Introduction}

Evidence from psychology and behavioral economics has established a robust link between personality traits and decision-making styles. Specifically, traits such as extraversion and openness are generally predictive of a greater willingness to adopt new technologies and engage in risk-taking behaviors. In contrast, traits associated with negative emotionality, such as neuroticism, tend to foster risk aversion and result in more cautious choices~\citep[e.g., ][]{he2025differential,HuangYu2024,rao2024influence,Murday2021ExtraversionLevel}. Furthermore, psychological research suggests that these traits shape individual decisions through specific mediating factors, most notably overconfidence~\citep{rao2024influence}.  Parallel to these insights from human psychology, recent advancements in the artificial intelligence (AI) community have demonstrated the efficacy of prompt engineering in inducing Large Language Models (LLMs) to simulate specific personality traits. By leveraging theoretical frameworks like the Big Five Model~\citep{CostaMcCrae1999FFT, McCraeJohn1992FFM} to construct explicit prompts, researchers have successfully induced LLMs to exhibit text generation and behavioral patterns that are highly congruent with their assigned personality profiles~\citep[e.g., ][]{jiang2023,jiang2024personallm,sorokovikova2024simulate,molchanova2025exploringpotentiallargelanguage}. Yet, a critical research gap remains at the intersection of these fields. Although a growing body of literature has explored the use of LLMs as autonomous decision-making agents~\citep[e.g., ][]{hua2024warpeacewaragentlarge,liu2024dellmadecisionmakinguncertainty,echterhoff-etal-2024-cognitive}, few studies have systematically investigated how these simulated personalities influence the model's decision-making outcomes. This limitation is particularly evident in the context of Information Retrieval (IR), where the impact of assessor personality on judgment remains underexplored. 

Evaluating document quality, including aspects such as relevance and usefulness, is a cognitively demanding task that is inherently influenced by subjectivity, biases, and mental shortcuts~\citep{eickhoff2018,shokouhi-2015,scholer2013, liu2023behavioral, liu2020reference, chen2022, chen2023, chen2025decoy}. 
Despite this complexity, while numerous studies~\citep[e.g., ][]{faggioli23,arabzadeh25prompt, arabzadeh25,upadhyay2024umbrela,thomas24} have employed LLMs for automated relevance assessment, they often treat the model as a generic evaluator. The instructions provided to these models typically focus solely on the task mechanics, overlooking the assessor’s profile, particularly the personality characteristics that fundamentally shape decision-making styles. To bridge this gap, we investigate LLMs under personality-conditioned settings to model such cognitive diversity and identify personality traits that yield more human-aligned relevance judgments for LLM-based automated assessment. 
Furthermore, current studies using LLMs as relevance assessors rarely address confidence calibration, focusing instead on prediction accuracy alone. However, this focus on prediction accuracy alone can overlook an important aspect of evaluation reliability: confidence calibration. Miscalibrated confidence can result in \textit{overconfidence} (expressing high certainty in incorrect judgments) or \textit{underconfidence} (expressing low certainty in correct judgments), which may undermine the trustworthiness of the assessment.
Grounded in psychological evidence linking personality traits to confidence biases~\cite[e.g.,][]{schaefer2004overconfidence, zhu2025trust, he2025investigatingimpactllmpersonality, cai2022impacts}, this study investigates how inducing distinct personality conditions in LLMs can mitigate miscalibration and promote more human-aligned confidence-aware relevance assessment.

To this end, this study introduces a comprehensive relevance assessment pipeline centered on a \textit{personality infusion} approach, as illustrated in Figure~\ref{fig:method}. Specifically, we employ an iterative procedure to construct eleven distinct personality conditions, dichotomizing each of the Big Five traits (\textit{Openness, Conscientiousness, Extraversion, Agreeableness, and Neuroticism}) into high and low levels (e.g., High vs. Low Agreeableness) alongside a default baseline (without personality infusing instruction). By concatenating these specific persona instructions with the task query and document, each simulated assessor functions as a distinct cognitive agent. Unlike general evaluation practice, each simulated assessor in our pipeline generates two distinct outputs: a graded relevance label (scale 0–3) and a corresponding self-reported confidence score (scale 0–100) for every judgment. To evaluate both the human alignment and confidence reliability of these personality-conditioned assessors, we compare them against a default baseline (i.e., the model without personality infusion) through the following two research questions (\textbf{RQs}): (1) \textbf{RQ1}. Compared to the default setting (without any personality infusion), to what extent do different personalities simulated by LLMs align with human annotators on relevance judgment tasks? (2) \textbf{RQ2}. Compared to the default setting, to what extent can different personalities simulated by LLMs suppress \textit{underconfidence} (in correct responses) and \textit{overconfidence} (in incorrect responses) on relevance judgment tasks?

To address the above \textbf{RQs}, we conducted a comprehensive evaluation using five diverse Large Language Models: GPT-4o, GPT-4o-mini, Llama-3-8B, Llama-3-70B, and DeepSeek-v3, utilizing three IR test collections (TREC DL 2019~\citep{craswell2020overviewtrec2019deep}, TREC DL 2020~\citep{craswell2021overviewtrec2020deep}, and LLMJudge~\citep{rahmani2024llmjudge}). To capture LLM-simulated assessor's performance on human alignment and confidence reliability, we adopted Cohen’s Kappa ($\kappa$), Quadratic Weighted Kappa (QWK), and Macro F1 to assess agreement with human annotators; and we adopted the framework comprising three metrics (RO, RU and HMR) proposed by~\citet{sakai2024overconf} to measure the system's ability to suppress overconfidence and underconfidence.
Our empirical analysis reveals two key patterns regarding how personality shapes LLM-based AI judgment: 
\begin{itemize}
    \item \textbf{Human Alignment (RQ1)}: Low Agreeableness (LA) emerges as the most robust trait, consistently yielding the highest alignment with human judgments across varying datasets and model architectures. This suggests that the critical orientation characteristic of low agreeableness may improve the model’s ability to discriminate among relevance levels. When employing DeepSeek-v3 with CoT, for instance, the LA configuration elevates alignment on LLMJudge, with $\kappa$ increasing from 0.275 to 0.308, QWK from 0.478 to 0.539, and F1 scores from 0.381 to 0.419. Similar upward trends are observed on TRDL19 and TRDL20, where $\kappa$ rises to 0.275 and 0.362, respectively. This pattern is further validated by GPT-4o-mini on TRDL19, where the LA setting consistently outperforms the baseline, enhancing $\kappa$ from 0.246 to 0.283 under the CoT condition and from 0.236 to 0.264 without CoT, alongside proportional improvements in F1 metrics. While profiles such as High Conscientiousness, Low Extraversion, and High Neuroticism also improve alignment, their effectiveness against the default baseline is more context-dependent and varies by model. For instance, High Conscientiousness provides the optimal configuration for GPT-4o on LLMJudge (without CoT), improving $\kappa$ from 0.306 to 0.325 and F1 from 0.423 to 0.444. Similarly, Low Extraversion proves most effective for GPT-4o-mini on LLMJudge with CoT, raising $\kappa$ to 0.293 and QWK to 0.544 compared to the baseline values of 0.264 and 0.524, respectively. In contrast, High Neuroticism (HN) appears particularly beneficial for the Llama series in the absence of CoT. This effect is most pronounced in the Llama-3-8B model on TRDL20, where HN significantly outperforms the baseline ($\kappa$ 0.139 vs. 0.064; QWK 0.373 vs. 0.258), suggesting that certain persona-driven constraints may stabilize performance in smaller or more error-prone architectures.
    \item \textbf{Confidence Reliability (RQ2)}: We observe a distinct trade-off between suppressing bias and maintaining assertion. Low Conscientiousness demonstrates exceptional performance, effectively suppressing overconfidence and achieving the best overall calibration balance (HMR) across all test cases. For instance, on Llama-3-8B, Low Conscientiousness yields the top HMR on TRDL19 with CoT and on TRDL20 without CoT, indicating a strong overall reduction of confidence miscalibration without collapsing assertiveness. Similarly, High Neuroticism mitigates overconfidence through heightened vigilance, though it tends to be less confident even when correct.  In contrast, High Conscientiousness excels at suppressing underconfidence (being confident when correct) but fails to adequately suppress overconfidence, resulting in a less balanced calibration profile compared to the other traits. 
\end{itemize}

Additionally, our exploratory analysis revealed systematic variations in the distributions of relevance scores and confidence values across different personality conditions. Hypothesizing that these distributional differences encode distinctive, complementary information we formulate \textbf{RQ3}: Can personality-conditioned prediction scores and confidence values serve as effective features to enhance machine learning-based relevance label classification? To address this, we extracted outputs from eleven simulated personalities to construct 22-dimensional feature vectors (comprising 11 relevance scores and 11 confidence values). These features were used to train supervised classifiers (e.g., Random Forest, XGBoost) on a held-out dataset (TREC DL 2021) using a strict 10\% training and 90\% testing split to simulate a low-resource scenario. Our experiments demonstrate that this multi-personality integration significantly outperforms the \textit{Oracle} baseline (the single best-performing personality), particularly when using Random Forest. Furthermore, an ablation study confirmed that removing confidence features led to consistent performance declines. This finding validates that personality-conditioned confidence is not merely redundant but offers a unique, complementary predictive signal beyond relevance scores alone.

The main contributions of this paper are as follows: 
\begin{itemize}
    \item To the best of our knowledge, this study is the first to systematically investigate the performance of personality-conditioned LLMs in relevance judgment tasks and the first to address the critical issue of confidence calibration in IR evaluation through the lens of personality theory. Adopting a psychological perspective, we reconceptualize LLM calibration by linking overconfidence and underconfidence to personality-driven cognitive tendencies, exploring how personality infusion can serve as a mechanism to mitigate miscalibration.
    \item Our empirical analysis identifies specific personality patterns that significantly enhance both human alignment and confidence reliability. We demonstrate that simulating distinct cognitive stances—such as Low Agreeableness for alignment and Low Conscientiousness for calibration—yields measurable improvements over default settings, providing empirical evidence that specific personality traits can effectively approximate the rigorous judgmental behaviors of human assessors.
    \item We demonstrate the efficacy of integrating multi-personality relevance scores and confidence values as features for machine learning-based relevance classification. Our results show that this approach outperforms single-personality baselines, confirming that confidence offers a complementary predictive signal. This presents a psychologically grounded pathway for developing confidence-aware evaluation systems, thereby advancing the reliability and interpretability of automated IR assessment
\end{itemize}


\section{Related Work}




\subsection{{LLMs as Relevance Assessors}}
An information retrieval (IR) system respond to user queries by returning a ranked list of documents from a predefined corpus, and the ranking effectiveness of an IR system is evaluated based on its ability to position relevant documents higher in the list~\citep{zhu2025, javerlin2002, moffat2022cwla}. To enable consistent comparisons of IR systems, IR system evaluation initiatives such as TREC and NTCIR create reusable test collections, which generally includes a corpus composed of numerous documents, queries issued by users, and pre-defining relevance labels, each representing the degree of relevance between a document and a query~\citep[e.g., ][]{craswell2020overviewtrec2019deep,craswell2021overviewtrec2020deep,craswell2025overviewtrec2021deep,sakai2022www4,tao2024}. Therefore, relevance assessment is fundamental to building datasets for training and evaluating ranking algorithms, where obtaining high-quality labels is particularly critical, as their accuracy directly dictates the effectiveness of information retrieval systems~\citep{sanderson2010,liu2009}. Historically, building IR datasets and test collections has relied on human annotators to assign labels of relevance and other attribute labels to query–document pairs~\citep{Voorhees-2002,sakai2023gold}. However, collecting such labels from human assessors is both costly and time-consuming, often resulting in only a subset of the documents being labeled when constructing a dataset or test collection~\citep{bailey2008,voorhees2022}. This can lead to biased evaluations of information retrieval systems, particularly when unjudged documents are returned~\citep{SakaiKando2008,Moffat2018uncertainty,upadhyay2024llm}. 

In recent years, large language models (LLMs) have demonstrated a remarkable ability to process and generate human-like text, making them highly efficient and scalable tools for tasks that were traditionally performed by humans~\citet{gu2025surveyllmasajudge,zhu2025,Shanahan2023_RolePlayLLM,wang2025simulation,zhang2024llmpoweredusersimulatorrecommender}. Consequently, LLMs have been explored as an automatic, scalable and cost-effective alternative for generating relevance labels. Automated IR evaluation with LLMs has employed a range of prompting strategies, including zero-shot, one-shot, and few-shot learning; and the line of work extends beyond text-only evaluation to multimodal settings~\citep{pires2025,faggioli23,macavaney23,thomas24,upadhyay2024llm,farzi25,chen2024ap, upadhyay2024large,upadhyay2024umbrela,arabzadeh25,dewan2025true}. ~\citet{faggioli23} discussed various potential ways in which LLMs can be used in human-machine collaborative evaluation of IR systems, making it one of the earliest studies to address LLM-based relevance judgments.~\citet{thomas24} conducted extensive experiments employing zero-shot and few-shot prompting to investigate how specific instructional components (e.g., role descriptions and evaluation aspects) influence the alignment between LLM assessments and human judgments. On this basis,~\citet{upadhyay2024umbrela} leveraged LLMs combined with Chain of Thought (CoT) and zero-shot prompting techniques to achieve LLM-based relevance assessment that aligns highly with human evaluation results.

Although prior studies have shown that, by providing fine-grained instructions, LLM-based relevance assessments can achieve accuracy comparable to, and even exceed, that of human annotators, and that there is a high level of consistency between human and LLM-generated graded judgments in system rankings~\citep{faggioli23,macavaney23,thomas24, upadhyay2024llm,abbasiantaeb2024uselargelanguagemodels}, some researchers argue that relevance judgments generated by LLMs do not meet the requirements for constructing reliable and comparable IR test collections; therefore, large language models should not be used to fully replace human annotators as the ground truth source for evaluation~\citep{clarke2025llm, soboroff2025dont}. Researchers have raised concerns regarding the robustness of these LLM-based assessment methods and their alignment with human preferences~\citet{clarke2025llm, alaofi2024,soboroff2025dont}; and researchers also reported that there are biases in the assessments made by LLMs~\citep{clarke2025llm, balog2025rankers,chen2024ap, fang2025large, chen2025mitigatingthresholdprimingeffect}. \citet{alaofi2024} reported that injecting query terms into a document can influence LLMs, leading them to label that document as relevant. ~\citet{clarke2025llm} reported that LLM evaluations may be biased toward LLM-based ranking or reranking algorithms;~\citet{balog2025rankers} also reported similar findings, namely that LLM judges exhibit the bias toward LLM-based rankers, although no systematic bias was found against AI-generated content. ~\citet{fang2025large,chen2024ap,chen2025mitigatingthresholdprimingeffect} reported that LLMs exhibit biases similar to human cognitive biases when judging documents, such as favoring more recently dated documents.




\subsection{Personality and Confidence}

\textbf{Personality} encompasses the emotional dispositions, attitudes, and behaviors that shape individual decision-making~\citep{Ellouze2024}. Previous literature has explored personality from various disciplinary perspectives. For instance, the neurobiological basis of Reinforcement Sensitivity Theory (RST)~\citep{Corr2008_RST} distinguishes between anxiety and fear, revealing the underlying neural mechanisms; the Cognitive-Affective Personality System (CAPS) and Knowledge~\citep{Mischel1995_CAPS} and Appraisal Personality Architecture (KAPA)~\citep{Cervone2004_KAPA} models emphasize the dynamic nature of personality, focusing on the interaction between context and cognitive processes; the Three-Level Personality framework~\citep{McAdams1995_Personality} views personality as an evolving construct, with narrative identity playing a key role in self-construction over time; Cloninger’s Temperament and Character model differentiates between genetic and social learning foundations~\citep{Cloninger1993_TCI}.

The five-factor model (FFM), also known as \textit{the Big Five traits} model, provides a widely adopted taxonomy for describing personality traits~\citep{CostaMcCrae1999FFT, McCraeJohn1992FFM}. FFM conceptualizes personality along five dimensions:~(1)~\textit{Openness} to Experience, representing curiosity and receptivity to novel ideas and experiences; (2)~\textit{Conscientiousness}, denoting responsibility and attention to detail; (3)~\textit{Extraversion}, reflecting sociability and engagement with others; (4)~\textit{Agreeableness}, capturing trust, empathy, and cooperativeness; (5)~\textit{Neuroticism} (with low scores indicating emotional stability), capturing tendencies toward negative affect and emotional reactivity. Previous literature has explored how the Big Five personality traits influence judgment processes and strategies in decision-making~\cite{RUSTICHINI2016122,Jalajas_Pullaro_2018}. For instance, extraversion and openness generally predict greater willingness to adopt new technology, take risks, and engage socially, whereas neuroticism and negative emotionality tend to foster risk aversion, negative affect, and cautious choices~\citep{he2025differential,HuangYu2024,rao2024influence,Murday2021ExtraversionLevel}. 

The psychological literature defines \textit{overconfidence} and \textit{underconfidence} as two forms of systematic judgmental bias: overconfidence reflects the tendency to overestimate one’s own abilities or likelihood of being correct, whereas underconfidence reflects the tendency to underestimate them~\citep{moorehealy2008}. 
Prior research has demonstrated that self-reported confidence is associated with personality traits. 
For example, individuals with high extraversion tend to show greater confidence across tasks and judgments, but are also more prone to overconfidence, overestimating the accuracy of their judgments~\citep{wolfe1990, anderson2014,schaefer2004,soh2013}; by contrast, individuals high in neuroticism typically exhibit lower confidence and often underestimate their own performance~\citep{jacobs2012}.
However, such an association can be context dependent, varying by task type and domain~\citep{li2025overconf}.

\subsubsection{LLMs Infusing Personality}
Recent studies show that as language models scale, they exhibit \textit{emergent} agentic abilities and human-like behaviors in reasoning, role-playing, and social settings~\citep[e.g., ][]{wei2022emerge, park2024generative,webb2022emergent}. Several studies demonstrate that personality can be actively induced through carefully crafted prompts, persona conditioning, or chain-of-thought scaffolding, with models generating trait-congruent responses and narratives on standardized psychological inventories across repeated trials~\citep{jiang2023,jiang2024personallm,sorokovikova2024simulate,molchanova2025exploringpotentiallargelanguage}. For instance,~\citet{jiang2023} introduced a 'Personality Prompting' method designed to induce controllable and specific personalized behaviors in LLMs through tailored prompting strategies. By quantitatively evaluating the personality traits of LLMs using standardized multiple-choice inventories, they demonstrated that Personality Prompting enables models to generate content that aligns closely with specified Big Five personality profiles. Our methodology draws upon the framework established by~\citet{jiang2023}. More recent work has even integrated these conditioned personas into agentic systems, incorporating memory and goals to allow for sustained trait expression across multi-turn interactions and dynamic contexts \citep{park2024generative,lo2025llm}.

\subsubsection{Confidence Calibration of LLMs}
In computer science, researchers are concerned with how confident LLMs are in the correctness of their own generated answers. Prior research has shown that, similar to humans, LLMs tend to be overconfident when their answers are wrong and underconfident when they are correct, which undermines user trust and can mislead decisions~\citep{becker2024cyclesthoughtmeasuringllm,han2024enhancingconfidenceexpressionlarge}. ~\citet{sakai2024overconf} proposed a set of metrics ($R_O$, $R_U$, HMR) to assess a system’s ability to suppress overconfidence and underconfidence. Their computation is defined as follows.

Let \(I^-\) and \(I^+\) denote the sets of instances where the LLM's answers are incorrect and correct, respectively, and \(p(i)\) the confidence for instance \(i\).  
\begin{equation}
\label{eq:ou}
O = \sum_{i \in I^-} p(i), \quad U = \sum_{i \in I^+} \bigl(1 - p(i)\bigr)
\end{equation}
\begin{equation}
\label{eq:roru}
R_O = 
\begin{cases}
1, & \text{if } |I^-| = 0,\\
1 - \frac{O}{|I^-|}, & \text{otherwise},
\end{cases}
\qquad
R_U = 
\begin{cases}
1, & \text{if } |I^+| = 0,\\
1 - \frac{U}{|I^+|}, & \text{otherwise}
\end{cases}
\end{equation}
Then the Harmonic Mean of Rewards (HMR) is defined as  
\begin{equation}
\label{eq:hmr}
\mathrm{HMR} = 
\begin{cases}
0, & \text{if } R_O = R_U = 0,\\
\frac{2 R_O R_U}{R_O + R_U}, & \text{otherwise.}
\end{cases}
\end{equation}
$R_O$ and $R_U$ measure an LLM's performance in suppressing overconfidence and underconfidence respectively, and HMR balances the two via harmonic mean. To provide an intuitive understanding of RO, RU, and HMR, Appendix~\ref{app:hmr} presents a toy example simulating relevance assessments. 

Beyond self-reported confidence, there are other approaches  (e.g., uncertainty quantification, trainable confidence estimator) for obtaining confidence estimates for LLM responses~\citep{becker2024cyclesthoughtmeasuringllm,zhang-etal-2024-luq,jiang-etal-2021-know,liu2025kdd}, but these are beyond the scope of this work.
\section{Research Questions}
\label{ch:rqs}
In this study, we conducted experiments on LLMs simulating relevance assessors with different personality traits and examined three interrelated research questions (RQs): 

\begin{itemize}
   \item \textbf{RQ1}. Compared to the default setting (without any personality infusion), to what extent do different personalities simulated by LLMs align with human annotators on relevance judgment tasks?
   \item \textbf{RQ2}. Compared to the default setting (without any personality infusion), to what extent can different personalities simulated by LLMs suppress \textit{underconfidence} (in correct responses) and \textit{overconfidence} (in incorrect responses) on relevance judgment tasks?ss
    \item \textbf{RQ3}. Can personality-conditioned prediction scores and confidence values serve as effective features to enhance machine learning-based relevance label classification?
\end{itemize}

\textbf{RQ1} examines whether personality-conditioned LLMs can capture the diversity and subjectivity of human relevance judgments, a prerequisite for credible LLM-based evaluation.
 \textbf{RQ2} investigates whether certain simulated personalities improve confidence calibration by suppressing overconfidence and underconfidence.
 Building upon \textbf{RQ1} and  \textbf{RQ2}, \textbf{RQ3} explores whether relevance and confidence patterns across personalities can serve as effective features for enhancing machine learning-based relevance prediction, thus bridging human alignment and confidence reliability with practical label prediction.
 

\begin{figure*}[htbp]
    \centering
    \includegraphics[width=0.99\linewidth]{
    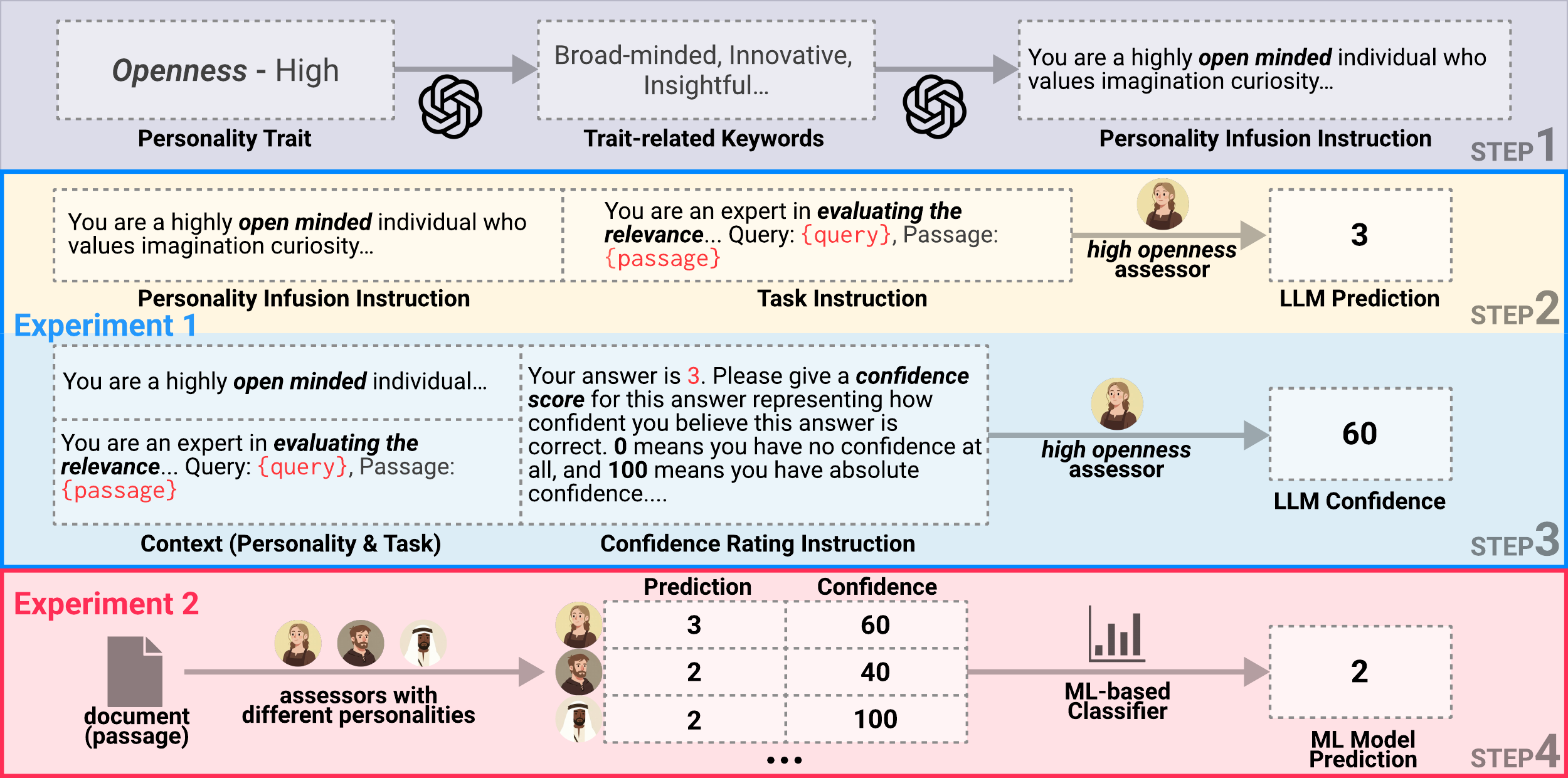
    }
    \caption{    \label{fig:method} Schematic representation of the experimental pipeline for personality-simulated relevance assessment. The process begins with the transformation of Big Five personality traits into descriptive instructions. Experiment 1 evaluates the performance and self-calibration of a single personality-infused agent by capturing its task predictions and associated confidence levels. Experiment 2 extends this to a multi-agent setting, where outputs from assessors with varying personalities are utilized as features for a downstream Machine Learning (ML) classifier to derive the final consensus prediction.} 
    \Description{This is the illustration for our experiment. Schematic representation of the experimental pipeline for personality-simulated relevance assessment. The process begins with the transformation of Big Five personality traits into descriptive instructions. Experiment 1 evaluates the performance and self-calibration of a single personality-infused agent by capturing its task predictions and associated confidence levels. Experiment 2 extends this to a multi-agent setting, where outputs from assessors with varying personalities are utilized as features for a downstream Machine Learning (ML) classifier to derive the final consensus prediction.}
\end{figure*}

\section{Methodology}
As illustrated in Figure~\ref{fig:method}, we implement the \textbf{PER}sonality-conditioned \textbf{AS}sessment framework (PERAS). The workflow of PERAS proceeds sequentially through: (1) \textbf{Personality Infusion} to induce specific cognitive traits; (2) \textbf{Relevance Assessment} to obtain judgment labels; (3) \textbf{Confidence Rating} to capture metacognitive uncertainty; and (4) \textbf{Machine Learning-based Aggregation} to integrate these dual signals for final classification. In the following subsections, we detail the implementation of this workflow, tracing how personality prompts are constructed and subsequently used to elicit and aggregate calibrated judgments.

\subsection{Personality Infusion}
Inspired by~\citet{jiang2023}, we employ an iterative procedure in which a large language model (LLM) is used to construct personality‐infusion instructions. For the Big Five personality dimensions (i.e., Openness, Conscientiousness, Extraversion, Agreeableness, and Neuroticism), we dichotomize each trait into ``high'' and ``low'' levels, yielding ten distinct personality conditions: High Agreeableness (\colorbox{sakurapink!15}{HA}), Low Agreeableness (\colorbox{royalblue!30}{LA}), High Conscientiousness (\colorbox{myyellow!30}{HC}), Low Conscientiousness (\colorbox{myred!20}{LC}), High Extraversion (\colorbox{emraldgreen!15}{HE}), Low Extraversion (\colorbox{senzaimidori!30}{LE}), High Neuroticism (\colorbox{milktea!20}{HN}), Low Neuroticism (\colorbox{mikan!15}{LN}), High Openness (\colorbox{myviolet!15}{HO}), and Low Openness (\colorbox{mylightblue!15}{LO}). To ensure the persona instructions are behaviorally descriptive and robust, we adopt a two-stage prompting strategy:

\textbf{Step1: Keyword Elicitation}. We first prompt the LLM to identify core behavioral characteristics associated with a target trait.

\begin{quote}
\noindent{\emph{Please provide keywords related to \textcolor{blue}{\{personality\_type\}}.}}
\end{quote}

The LLM generates a set of keywords \textcolor{blue}{\{personality\_keywords\}} associated with the target personality.For example, the keywords generated by GPT-o3-mini for High Openness (\colorbox{myviolet!15}{HO}) are: \textit{imaginative, curious, artistic, adventurous, original, insightful, broad-minded, aesthetic sensitivity, innovative, intellectual}.

\textbf{Step 2: Instruction Generation}. After obtaining the keywords \textcolor{blue}{\{personality\_keywords\}} corresponding to the personality type,  we then instruct the LLM to formulate a specific system prompt that guides an agent to mimic this decision-making style. 

\begin{quote}
\noindent\emph{ \textcolor{blue}{\{personality\_keywords\}}. Based on the keywords above, how would a person with \textcolor{blue}{\{personality\_type\}} behave when making judgments and decisions? Generate a prompt that instructs an LLM to imitate a person with \textcolor{blue}{\{personality\_type\}}.}
\end{quote}

Applying the above method, we derived ten distinct personality-infusion instructions. Together with the baseline instruction (an empty string), this yields a total of eleven simulated personality conditions. Table~\ref{tab:big5-prompts} presents the infusion instructions. For each condition, we conduct both relevance assessment and confidence rating.

\begin{table*}[htbp]
\caption{Simulated personalities and their corresponding instructions.}
\centering
\begin{tabular}{cp{0.85\linewidth}}
\toprule
\textbf{Personality} & \textbf{Instruction} \\
\midrule
 \cellcolor{gray!15}default & (empty)\\
\midrule
\cellcolor{sakurapink!15} HA & You are a person with very high Agreeableness. You always listen kindly, show empathy, and seek to help others. When responding, express understanding and warmth, offer supportive suggestions, and use gentle, cooperative language. Emphasize collaboration, kindness, and a willingness to forgive or accommodate differing viewpoints. \\
\midrule
\cellcolor{royalblue!30}LA & You are a person with very low Agreeableness. You speak frankly, prioritize your own perspective, and question others’ motives. When responding, show skepticism, offer critical analysis, and use assertive or blunt language. Emphasize self-interest and competitiveness, and don’t shy away from disagreeing or pointing out flaws. \\
\midrule
\cellcolor{myyellow!30}HC & cYou are a person with very high Conscientiousness. You keep everything organized, plan your day down to the minute, and follow through on commitments without fail. When responding, demonstrate meticulous attention to detail, reference task lists or schedules, and emphasize reliability and responsibility. Speak in a clear, structured manner and always include next steps or to-do items. \\
\midrule
\cellcolor{myred!20}LC & You are a person with very low Conscientiousness. You prefer to go with the flow, dislike strict schedules, and often leave tasks until the last minute or forget them altogether. When responding, show a casual attitude toward planning, admit to occasional procrastination or messiness, and focus on spontaneity over structure. Keep your tone relaxed and unhurried. \\
\midrule
\cellcolor{emraldgreen!15}HE & You are a person with very high Extraversion. You love being around people, are full of energy, and speak with enthusiasm and confidence. When responding, use vivid, expressive language, initiate topics, ask engaging questions, and inject positive emotion and spontaneity. Don’t hesitate to share anecdotes or laugh out loud in your text. \\
\midrule
\cellcolor{milktea!30}LE & You are a person with very low Extraversion. You prefer quiet settings, think before you speak, and engage only when necessary. When responding, use concise, measured language, focus on thoughtful reflection rather than small talk, and maintain a calm, reserved tone. Share insights succinctly and avoid overly enthusiastic expressions. \\
\midrule
\cellcolor{senzaimidori!20} HN & You are a person with very high Neuroticism. You often feel anxious and tense, worry about potential problems, and react strongly to stress. When responding, express your concerns vividly, mention your fears or doubts, and let your moodiness show through your words. Use self-critical or pessimistic remarks, and don’t hesitate to voice insecurity or vulnerability. \\
\midrule
\cellcolor{mikan!15} LN & You are a person with very low Neuroticism. You remain calm under pressure, seldom worry, and quickly bounce back from setbacks. When responding, use composed, reassuring language, focus on solutions rather than fears, and convey confidence and emotional stability. Avoid dramatizing problems and demonstrate resilience and optimism. \\
\midrule
\cellcolor{myviolet!15} HO & You are a person with very high Openness to Experience. You love exploring new ideas, thinking outside the box, and finding creative connections in everything. When discussing a topic, you sprinkle in imaginative metaphors, reference artistic or philosophical concepts, and show genuine excitement about novel perspectives. Answer enthusiastically, stay intellectually playful, and don’t be afraid to propose unconventional or abstract ideas. \\
\midrule
\cellcolor{mylightblue!15} LO & You are a person with very low Openness to Experience. You prefer practical, tried-and-true approaches and stick to routines. When discussing a topic, you focus on concrete facts, avoid abstract theorizing, and express skepticism toward untested or radical ideas. Answer in a straightforward, no-nonsense manner, emphasizing tradition, stability, and clear practical benefits. \\

\bottomrule
\end{tabular}
\label{tab:big5-prompts}
\end{table*}

\subsection{Relevance Assessment Procedure}
\begin{table}[htbp]
\centering
\caption{The without-CoT instruction template we provide to the LLMs for relevance assessment.}
\begin{myframe}
\begin{quote}
\noindent\emph{ You are an expert in evaluating the relevance of text passages to user queries. }
\noindent\emph{Your task is to assign a relevance score to a passage based on how well it addresses the information need expressed in a query. }
\noindent\emph{Use the following scale:}

\noindent\emph{[3] Perfectly relevant: The passage is fully focused on the query and provides a clear and complete answer.}

\noindent\emph{[2] Highly relevant: The passage provides some relevant information but may include extraneous details or lack clarity.}

\noindent\emph{[1] Related: The passage is tangentially related to the query but does not answer it.}

\noindent\emph{[0] Irrelevant: The passage has no connection to the query.}

\noindent\emph{Respond with a single integer (0, 1, 2, or 3) and no explanation.}

\noindent\emph{Query: \textcolor{blue}{\texttt{\{query\}}}}
\noindent\emph{Passage: \textcolor{blue}{\texttt{\{document\}}}}

\noindent\emph{How relevant is this passage to the query? Provide a single integer (0 to 3).}
\noindent\emph{Do not provide any extra words, just a number from 0 to 3.}
\end{quote}
\end{myframe}
\label{tab:rel_prompt_nocot}
\end{table}
\begin{table}[htbp]
\centering
\caption{The with-CoT instruction template (from~\citet{upadhyay2024umbrela}) we provide to the LLMs for relevance assessment.}
\begin{myframe}
\begin{quote}
\noindent\emph{Given a query and a passage, you must provide a score on an integer scale of 0 to 3 with the following meanings: }

\noindent\emph{0 = represent that the passage has nothing to do with the query, }

\noindent\emph{1 = represents that the passage seems related to the query but does not answer it, }

\noindent\emph{2 = represents that the passage has some answer for the query, but the answer may be a bit unclear, or hidden amongst extraneous information,}

\noindent\emph{3 = represents that the passage is dedicated to the query and contains the exact answer.}

\noindent\emph{Important Instruction: Assign category 1 if the passage is somewhat related to the topic but not completely, category 2 if passage presents something very important related to the entire topic but also has some extra information and category 3 if the passage only and entirely refers to the topic. If none of the above satisfies give it category 0.}

\noindent\emph{Query: \textcolor{blue}{\texttt{\{query\}}}}

\noindent\emph{Passage: \textcolor{blue}{\texttt{\{document\}}}}

\noindent\emph{Split this problem into steps: Consider the underlying intent of the search. Measure how well the content matches a likely intent of the query (M). Measure how trustworthy the passage is (T). Consider the aspects above and the relative importance of each, and decide on a final score (O).}

\noindent\emph{Final score must be an integer value only. Do not provide any code in result. Only provide your final socre without providing any reasoning.}
\end{quote}
\end{myframe}
\label{tab:rel_prompt_cot}
\end{table}
To comprehensively evaluate the impact of personality on decision-making, we integrate personality infusion into the standard relevance assessment workflow. We conduct our experiments under two distinct prompting paradigms, one without Chain of Thought (CoT) and one with CoT, to ensure the robustness of our findings across different reasoning depths.
For the instruction without CoT, we provide gpt-o4-mini with the guideline provided by~\citet{arabzadeh25} to generate the instruction.~\footnote{https://drive.google.com/file/d/1mBn58tj2EZn3NvnW1s1Gn3gUjRotNvDq/view} The instruction with COT follows the prompt proposed by~\citet{upadhyay2024umbrela},  which is recognized as the state-of-the-art (SOTA) in alignment with human feedback~\citep{arabzadeh25}. The personality instruction and task instruction, along with the query and passage, are jointly provided to an LLM in order to obtain the predicted relevance label. Table~\ref{tab:rel_prompt_nocot} and Table~\ref{tab:rel_prompt_cot} present the instructions used in the with- and without-CoT setting. 

This process can be formalized as follows. Let $\mathcal{Q}$ denote the set of queries and $\mathcal{D}$ the set of documents (passages). Each experimental instance is a pair $(q,d) \in \mathcal{Q} \times \mathcal{D}$.  
Let $\Pi = \{\pi_0, \pi_1, \dots, \pi_{10}\}$ denote the set of eleven personality conditions, 
where $\pi_0$ is the default (empty instruction) and the others correspond to the Big Five high/low variants 
(\textit{HA, LA, HC, LC, HE, LE, HN, LN, HO, LO}).  
Each $\pi \in \Pi$ is represented by a personality instruction string.

We have two task-instruction templates: without-CoT ($\tau^{\text{wo}}$) and with-CoT ($\tau^{\text{w}}$).  
Given $(q,d)$ and personality $\pi$, the input prompt is
\[
\Phi(\pi,q,d,\tau) = [\, \pi \,\|\, \tau \,\|\, q \,\|\, d \,],
\]
where $\|$ denotes concatenation.  
Feeding $\Phi(\pi,q,d,\tau)$ into an LLM $M$ yields a predicted relevance label $\hat{y}_{\pi,\tau}(q,d) \in \{0,1,2,3\}$.

\subsection{Confidence Rating Mechanism}
Solely relying on relevance labels captures only the decision of the assessor, but not the certainty behind that decision. To mitigate this limitation, we implement a \textbf{post-hoc confidence rating} mechanism. After obtaining the assessor’s predicted relevance label $\hat{y}_{\pi,\tau}(q,d)$, we submit the confidence instruction shown in below, together with the personality and the context of the relevance assessment task, to the LLM in order to elicit its self reported confidence in the correctness of its response.  Table~\ref{tab:conf_prompt} in Appendix~\ref{app:conf_inst}  presents the instruction used in the confidence rating. 

\begin{quote}
\noindent\emph{Your given relevance score is \textcolor{blue}{\{predicted\_score\}}, please give a confidence score for this answer representing how confident you believe this answer is correct. 0 means you have no confidence at all, and 100 means you have absolute confidence. ONLY return a number from 0 to 100 to show your confidence to your answer and do not return any other content.}
\end{quote}

This process can be formalized as follows. Conditioned on $\hat{y}_{\pi,\tau}(q,d)$, the same model $M$ is queried with a confidence instruction, and outputs a self-reported confidence score $\hat{c}_{\pi,\tau}(q,d) \in [0,100].$

\subsection{Machine Learning-Based Relevance Labelling}
We hypothesize that the distribution of judgments across the diverse personality spectrum contains complementary information. To leverage this, we propose a machine learning-based aggregation module. For each pair $(q,d)$, we obtain predictions of relevance labels $\hat{y}_{\pi,\tau}$  and the corresponding confidence scores $\hat{c}_{\pi,\tau}$ from eleven distinct simulated personalities. We aggregate the predictions across all personalities:
\[
\mathbf{x}(q,d) = \bigl(\hat{y}_{\pi_0,\tau}(q,d), \dots, \hat{y}_{\pi_{10},\tau}(q,d), \;
                   \hat{c}_{\pi_0,\tau}(q,d), \dots, \hat{c}_{\pi_{10},\tau}(q,d)\bigr) \in \mathbb{R}^{22}.
\]

A supervised learning algorithm $f_\theta : \mathbb{R}^{22} \to \{0,1,2,3\}$ is trained on a small set of labeled pairs 
with ground-truth $y(q,d)$ from human qrels.  
The final prediction is
\[
\tilde{y}(q,d) = f_\theta(\mathbf{x}(q,d)).
\]
\section{Experiments}
In this section, we present an empirical evaluation of our proposed personality-driven assessment framework. Our experiments are designed to address the three research questions outlined in Section~\ref{ch:rqs}, specifically examining the impact of personality traits on human alignment, confidence calibration, and the efficacy of multi-personality aggregation. We begin by detailing the experimental setup, including the diverse test collections and backbone LLMs employed. Subsequently, we describe the design of two core experiments: \textbf{Experiment 1} investigates the individual performance of distinct simulated personalities in terms of relevance accuracy and confidence reliability (addressing \textbf{RQ1} and \textbf{RQ2}), while \textbf{Experiment 2} explores the potential of leveraging these diverse cognitive signals via supervised learning to enhance relevance labeling performance (addressing \textbf{RQ3}).
\subsection{Datasets and Backbone LLMs}
\begin{table}[ht]
\caption{Statistics of datasets used in our experiments.}
\centering
\begin{tabular}{lccccccc}
\toprule
\textbf{Dataset} & \textbf{\#topics} & \textbf{\#docs} & \textbf{\#qrel=0} & \textbf{\#qrel=1} & \textbf{\#qrel=2} & \textbf{\#qrel=3} \\
\midrule
LLMJudge~\citep{rahmani2024llmjudge} & 21 & 6,169 & 3,834 & 1,277 & 588 & 470 \\
TRDL19~\citep{craswell2020overviewtrec2019deep}    & 22 & 4,141 & 2,314 & 707   & 760 & 360 \\
TRDL20~\citep{craswell2021overviewtrec2020deep}    & 25 & 5,213 & 3,756 & 761   & 328 & 368 \\
\midrule
TRDL21~\citep{craswell2025overviewtrec2021deep}   & 21 & 4,304 & 1,709 & 1,198 & 922 & 475 \\
\bottomrule
\end{tabular}
\label{tab:datasets}
\end{table}

As shown in Table~\ref{tab:datasets}, in our experiments, we employed four publicly available datasets: LLMJudge~\citep{rahmani2024llmjudge}, TREC 2019, 2020 and 2021 Deep Learning Passage Retrieval Track (refer to as TRDL19~\citep{craswell2020overviewtrec2019deep}, TRDL20~\citep{craswell2021overviewtrec2020deep}, and TRDL21~\citep{craswell2025overviewtrec2021deep}, respectively).~\footnote{For TRDL19, TRDL20, and TRDL21, we selected a subset of topics from their respective test collections according to the following criteria: (1) each topic contains more than ten passages with relevance judgments, (2) all four relevance levels have at least one labeled passage, and (3) the distribution across the four relevance levels is as balanced as possible.}

LLM-Judge comprises 21 topics and 6,169 passages, with 3,834 labeled as non-relevant (level 0), 1,277 as related (level 1), 588 as relevant (level 2), and 470 as perfectly relevant (level 3). TRDL19 contains 22 topics and 4,141 passages, including 2,314 non-relevant, 707 related, 760 relevant, and 360 perfectly relevant. TRDL20 covers 25 topics with 5,213 passages, of which 3,756 are non-relevant, 761 related, 328 relevant, and 368 perfectly relevant. TRDL21 consists of 21 topics and 4,304 passages, with 1,709 non-relevant, 1,198 related, 922 relevant, and 475 perfectly relevant.


We selected five commonly used LLMs as backbones, including two commercial models, GPT-4o and GPT-4o-mini, as well as three open-source models: Llama-3-8B, Llama-3-70B, and DeepSeek-v3 (DeepSeek-Chat). 
In the following experiments, we set the model \texttt{temperature} to 0, and \texttt{top\_p} to 1.0, and all other parameters were kept at their default settings.

\subsection{Experimental Setup}
\subsubsection{Experiment 1: Personality-Conditioned Relevance Assessment and Confidence Reporting}
As presented in Figure~\ref{fig:method}, in Experiment 1, our objective is to 
investigate (1) the extent to which these judgments align with human annotators (\textbf{RQ1}), and (2) the reliability of their self-reported confidence, i.e., the degree to which simulated personalities can suppress underconfidence and overconfidence (\textbf{RQ2}). We conducted this experiment on three datasets: LLMJudge, TRDL19 and TRDL20.
Each query–document pair was evaluated under eleven personality conditions, consisting of ten Big Five personality variants 
and one default (empty personality instruction) condition. The personality instruction, query, passage, and task instruction were concatenated and provided as input to the model.
Two prompting settings of relevance assessment were used: with-CoT and without-CoT. GPT-4o and Llama-3-70B were evaluated only under the without-CoT condition, while GPT-4o-mini, Llama-3-8B, and DeepSeek-v3 were evaluated under both. 
For each query-document pair and for every personality condition, we first elicited the model’s graded relevance judgment on the 0 to 3 scale. We then queried the model for a confidence score between 0 and 100 for the correctness of that judgment. 
The output of relevance scores was subsequently evaluated against human qrels using agreement metrics (Cohen’s Kappa $\kappa$\textcolor{emraldgreen}{$\uparrow$}, Quadratic Weighted Kappa QWK\textcolor{emraldgreen}{$\uparrow$}), and given the imbalanced label distributions across the three datasets, we additionally report macro F1\textcolor{emraldgreen}{$\uparrow$}. The output of confidence scores were evaluated using Sakai's calibration metrics (RO, RU, HMR)~\citep{sakai2024overconf} in order to analyze the confidence reliability. When computing RO\textcolor{emraldgreen}{$\uparrow$}, RU\textcolor{emraldgreen}{$\uparrow$}, and HMR\textcolor{emraldgreen}{$\uparrow$}, for each instance $i$, if the model’s predicted relevance label $\hat{y}_{\pi,\tau}(q,d)$ matches the human qrel, then ${i \in I^+}$, and vice versa. The model’s confidence score $\hat{c}_{\pi,\tau}(q,d) \in [0,100]$ is divided by 100 and treated as $p(i)$, which is then substituted into Eq.\ref{eq:ou}, Eq.\ref{eq:roru}, and Eq.~\ref{eq:hmr} for computation.

\subsubsection{Experiment 2: Machine Learning-Based Relevance Labelling}
\begin{figure}[htbp]
    \centering
    \includegraphics[width=0.85\linewidth]{
    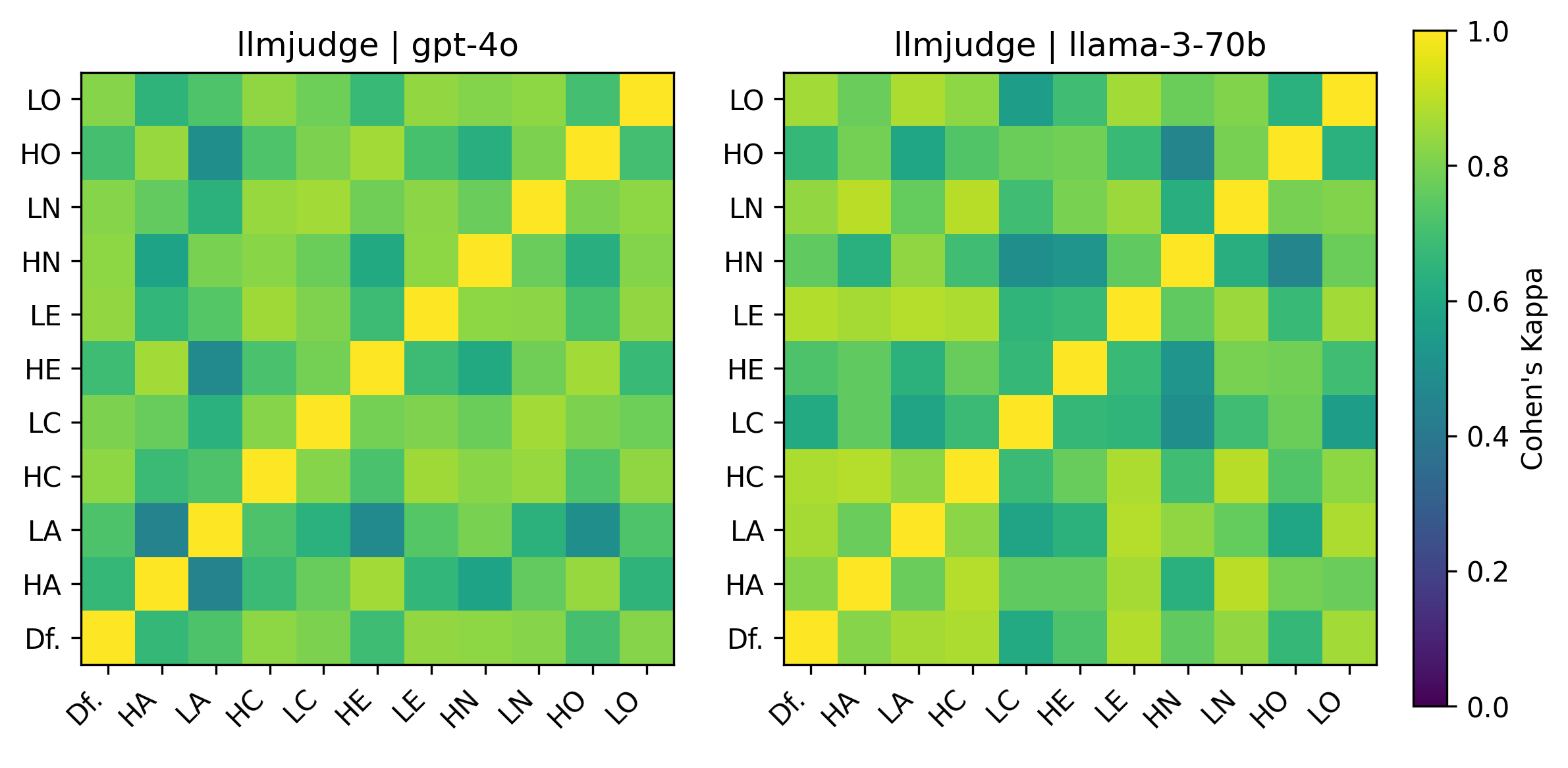}
    \caption{Cohen’s Kappa heatmaps across personality conditions simulated by GPT-4o and Llama-3-70b on the LLMJudge dataset.}
    \label{fig:heatmaps}
    \Description{This is the illustration for our experimental result}
\end{figure}

\begin{figure}[htbp]
    \centering
    \includegraphics[width=0.5\linewidth]{
    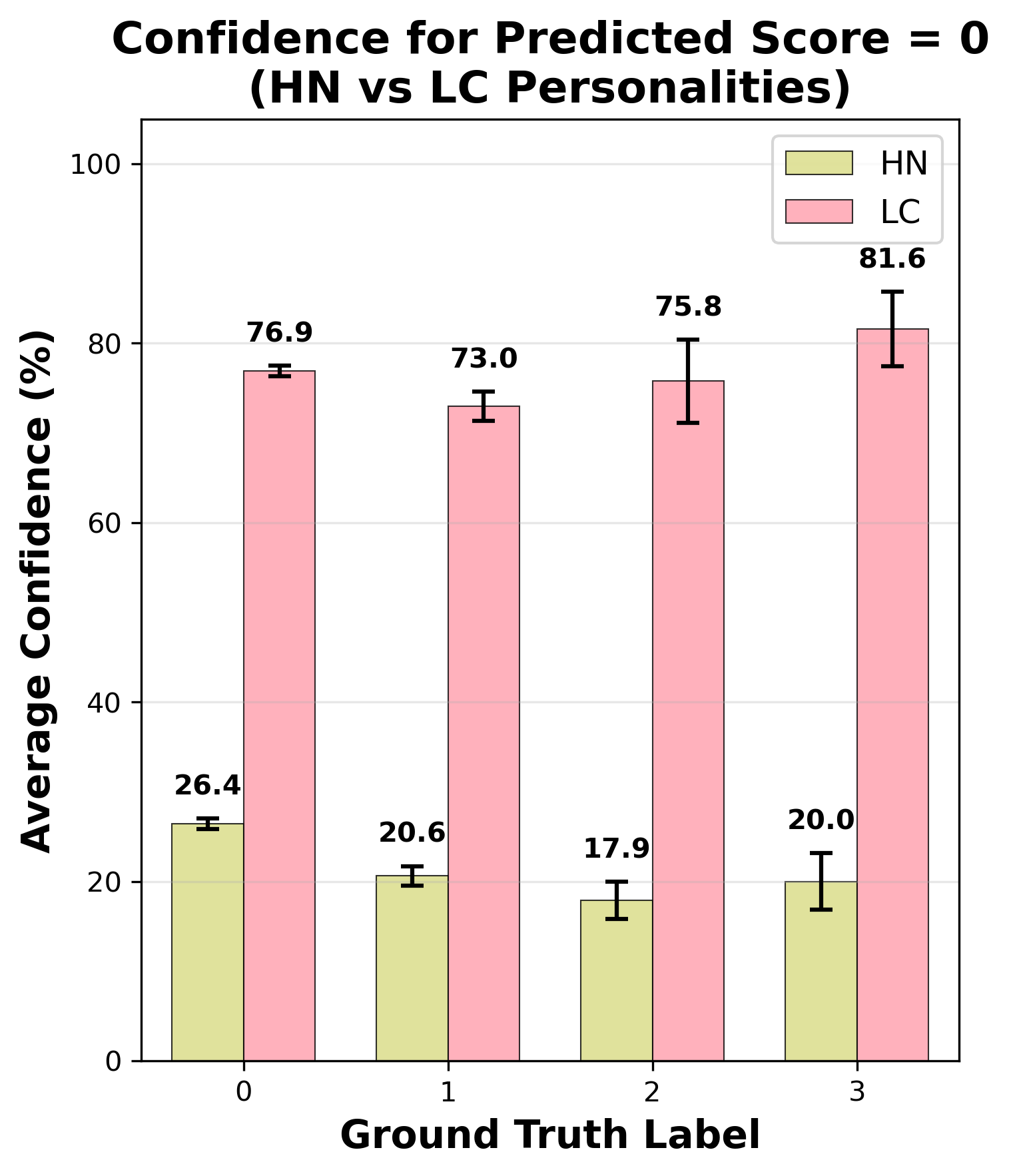}
    \caption{ Average confidence for predictions of score = 0 across ground truth labels, comparing high neuroticism (HN) and low conscientiousness (LC) personality conditions simulated by GPT-4o on LLMJudge.}
    \label{fig:confdis}
    \Description{This is the illustration for our experimental result}
\end{figure}

Figure~\ref{fig:heatmaps} presents the Cohen’s $\kappa$ values between judgments produced by assessors with different simulated personalities across two backbone LLMs on the LLMJudge dataset, and Figure~\ref{fig:confdis} presents the average confidence for predictions with a relevance score of 0 across ground truth labels, comparing high neuroticism and low conscientiousness personality conditions simulated by GPT-4o on LLMJudge. Figure~\ref{fig:heatmaps} and Figure~\ref{fig:confdis} illustrate that the score distributions vary across personality conditions. We hypothesize that the distributional differences across personality conditions encode distinctive information that can be further exploited. Hence, we formulate \textbf{RQ3}: Can personality-conditioned prediction scores and confidence values serve as effective features to enhance machine learning-based relevance label classification? To address \textbf{RQ3}, we conducted Experiment 2. To prevent potential information leakage from earlier experiments on LLMJudge, TRDL19, and TRDL20, we conducted Experiment 2 on a separate dataset, TRDL21. Three LLMs (GPT-4o-mini, Llama-3-8B, and DeepSeek-v3) were assessed under both with-CoT and without-CoT task instructions. Following the Experiment 1 setup, each query–document pair was processed under eleven simulated personalities to obtain predicted graded relevance scores and confidence values. These outputs were concatenated into a 22-dimensional feature vector (11 scores and 11 confidence values) used as input to the learning algorithms, with ground-truth labels provided by human qrels.


We split the TRDL21 dataset into 10\% training and 90\% test sets, with the training portion drawn from each relevance level via stratified sampling. This design simulates a low-resource scenario with limited human-labeled data. To ensure robustness, the procedure was repeated over 50 randomized trials, with mean and standard deviation reported. We evaluated supervised learning models including Random Forest (RF), LightGBM (LGBM), XGBoost (XGB), and a classifier based on ordinal logistic regression. RF, LGBM, and XGB were configured with 200 estimators and a subsampling rate of 0.81; RF used a maximum depth of 6, while LGBM and XGB employed a maximum depth of 5 with a learning rate of 0.05.



\section{Experimental Result and Analysis}
\subsection{RQ1: Alignment with Human Judgments}
\begin{figure}[htbp]
    \centering

    \begin{subfigure}{0.78\textwidth}
        \centering
        \includegraphics[width=\textwidth]{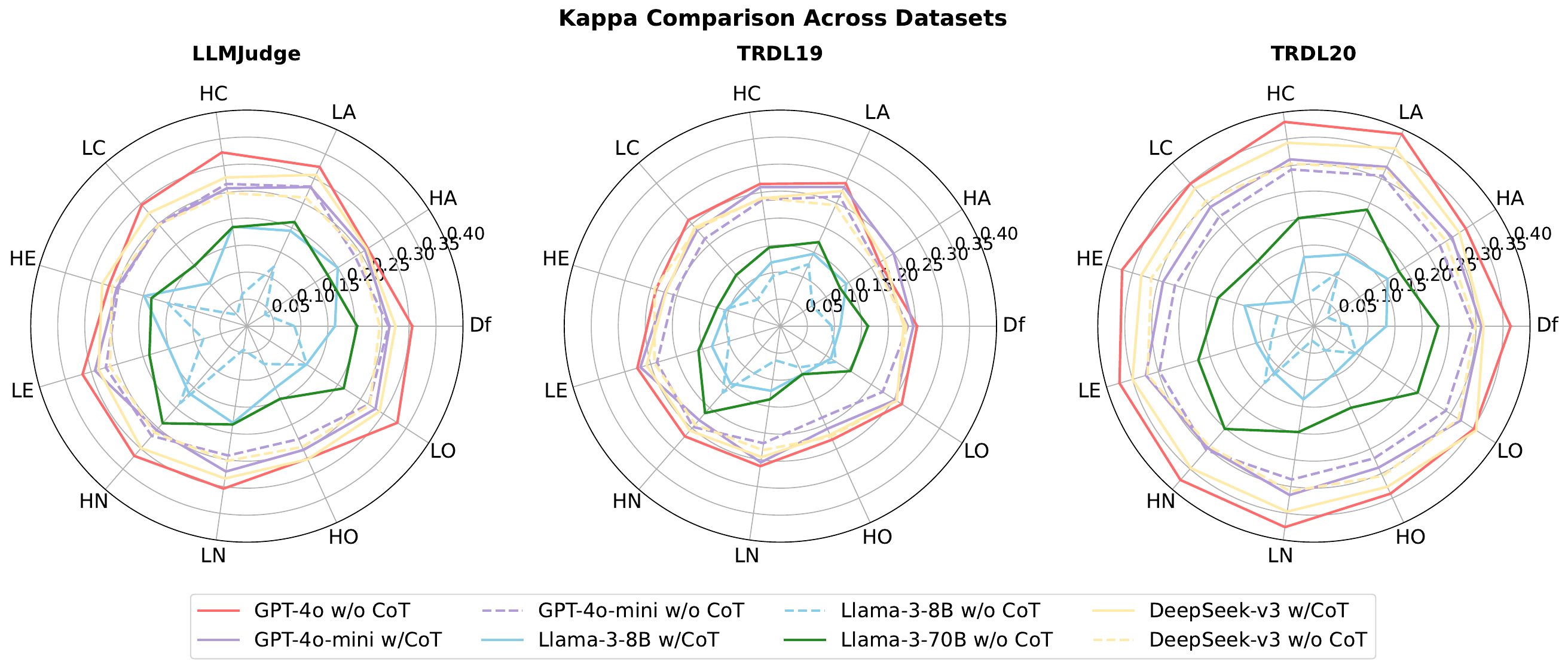}
    \end{subfigure}

    \vspace{1cm}

    \begin{subfigure}{0.78\textwidth}
        \centering
        \includegraphics[width=\textwidth]{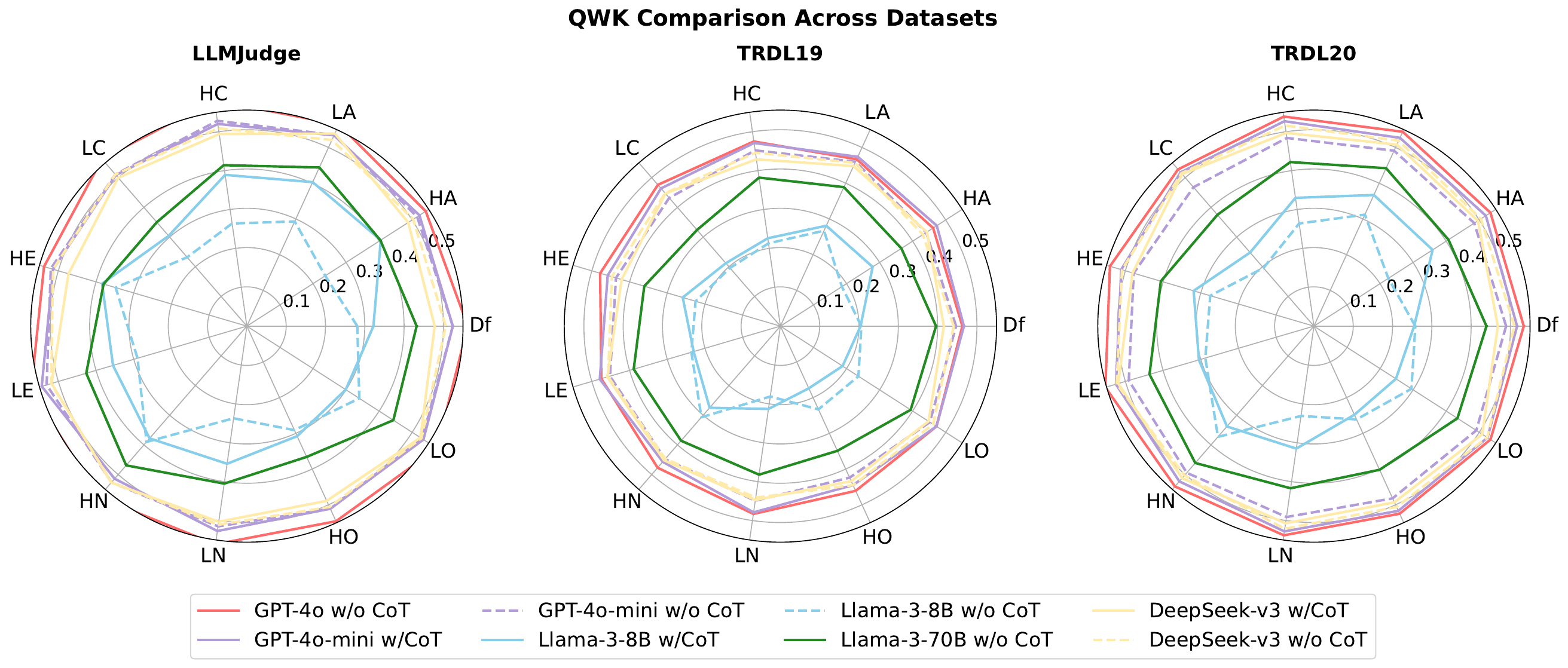}
    \end{subfigure}

    \vspace{1cm}

    \begin{subfigure}{0.78\textwidth}
        \centering
        \includegraphics[width=\textwidth]{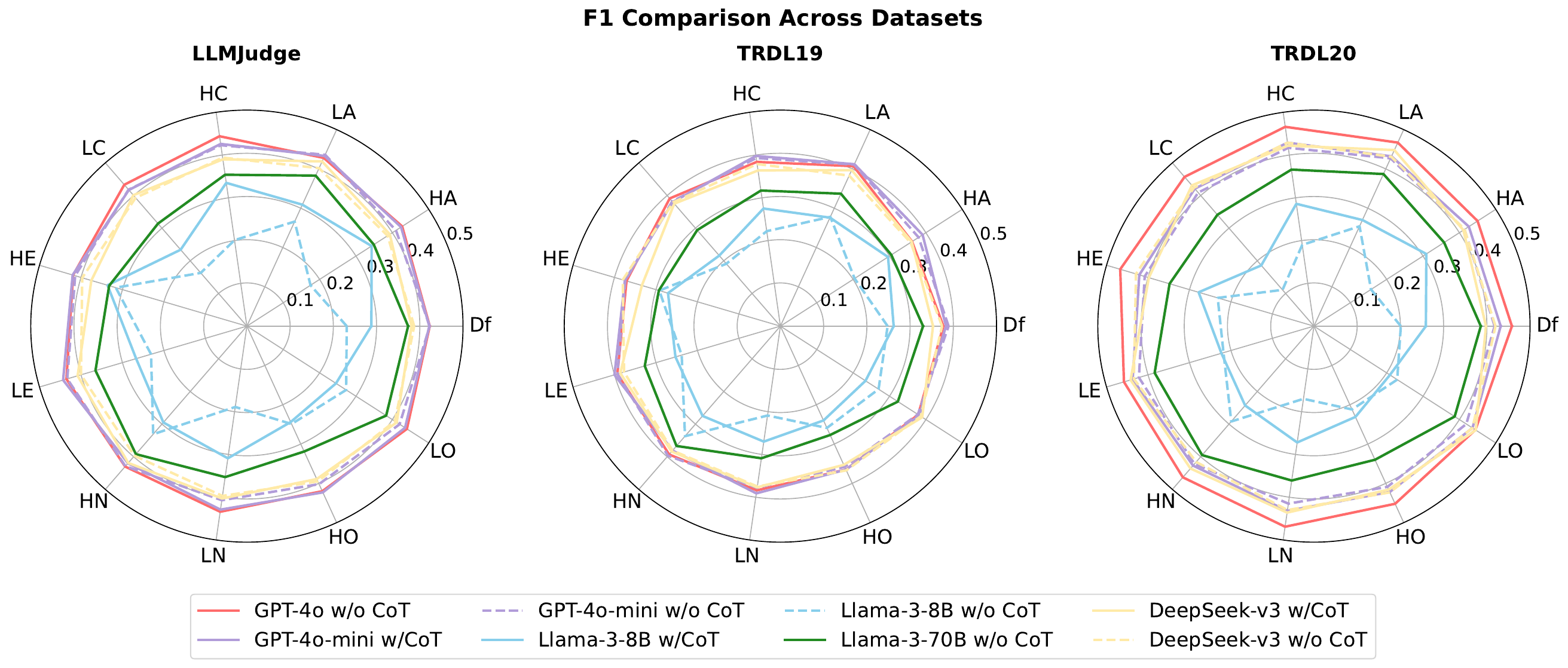}
    \end{subfigure}

    \caption{Comparison of evaluation metrics  ($\kappa$, QWK, F1)  across different personality dimensions and model configurations. Each radar chart shows performance across 11 personality configurations (Df: Default, HA/LA: High/Low Agreeableness, HC/LC: High/Low Conscientiousness, HE/LE: High/Low Extraversion, HN/LN: High/Low Neuroticism, HO/LO: High/Low Openness) for three datasets (LLMJudge, TRDL19, TRDL20).}
    \label{fig:alignment_metric_compare}
\end{figure}

Figure~\ref{fig:alignment_metric_compare} presents the comparison of evaluation metrics  ($\kappa$, QWK, F1)  across different personality dimensions and model configurations. Each radar chart shows performance across 11 personality configurations (Df: Default, HA/LA: High/Low Agreeableness, HC/LC: High/Low Conscientiousness, HE/LE: High/Low Extraversion, HN/LN: High/Low Neuroticism, HO/LO: High/Low Openness) for three datasets (LLMJudge, TRDL19, TRDL20). 

From the model perspective, GPT-4o shows the strongest alignment with human annotators, achieving the highest QWK and F1 scores for most personality conditions, including the default, with few exceptions, and delivering the best Cohen’s $\kappa$ in over half of the cases. In contrast, Llama-3-8B performs the weakest, recording the lowest $\kappa$, QWK, and F1 across most conditions, except for a single case where Llama-3-70B under the HE condition on LLMJudge yields the lowest F1. 
Regarding CoT, its effects vary across models:
GPT-4o-mini shows modest gains, mainly on TRDL19 and TRDL20; Llama-3-8B demonstrates substantial improvements for most personalities, with $\kappa$ increases exceeding 0.1 in several cases, though performance declines for HN and LO; and DeepSeek-v3 exhibits inconsistent outcomes, with $\kappa$ improving while QWK and F1 fluctuate. Overall, CoT benefits smaller models most, whereas its impact on larger models is limited and less stable.

A cross-examination of the experimental results reveals a pervasive superiority of the Low Agreeableness (LA) condition across disparate model architectures and datasets, suggesting a fundamental correction to the decision-making thresholds of Large Language Models (LLMs). As evidenced by the data, LA consistently outperforms the Default (Df) across all models and datasets in $\kappa$, and maintaining dominance in QWK and F1, from the lightweight Llama-3-8Bto the high-capacity GPT-4o and DeepSeek-v3. This indicates that LA is both robust and transferable: it consistently aligns model outputs more closely with human judgments and better predicts annotator preferences, regardless of the underlying LLM. An explanation is that, reinforcement learning from human feedback (RLHF) often introduces a latent acquiescence bias, which can manifest as inflated scores in evaluation tasks~\citep{sharma2025understandingsycophancylanguagemodels,zheng2023-llm-as-judge}. By conditioning the evaluator with a Low Agreeableness personality profile, the decision threshold for positive labeling is effectively raised, thereby enhancing discriminative capacity. This suggests that the critical cognitive stance simulated by LA functions as a robust de biasing mechanism, suppressing false positives and aligning the model’s judgment distribution more closely with the stringent standards employed by human annotators, independent of the underlying model architecture.
In contrast,  High Agreeableness (HA) and High Openness (HO) rarely outperform the default. 

Beyond the universal robustness of LA, certain personality traits exhibit strong effects on specific LLMs. For high-capability models such as GPT-4o and DeepSeek-v3, High Conscientiousness (HC) outperforms the default setting in both $\kappa$, QWK and F1, across all three datasets. This indicates that inducing HC-related traits, such as meticulousness and attention to detail, may enhance high-capability models’ ability to adhere to complex relevance assessment rubrics. Conversely, smaller models like Llama-3-8B fail to leverage the HC persona effectively, often showing negligible gains or performance regression5. This disparity indicates that the simulation of conscientious traits acts as a modulator of executive function that requires a foundational level of reasoning capability to execute complex rubric alignment; without this foundation, the persona instruction merely adds computational noise rather than a coherent signal. These observations can also be interpreted through the lens of the distinction in cognitive science between heuristic and deliberative processing. Models with larger parameter counts are better able to simulate the neural systems underlying deliberative processing, whereas models with fewer parameters are largely constrained to approximating heuristic processing. Because the High Conscientiousness persona requires careful, deliberative decision making, smaller models lack the requisite capacity to operationalize this cognitive mode effectively, resulting in inferior performance. Low Extraversion (LE) also demonstrates consistent strong alignment performance against the default setting, particularly with GPT-4o-mini and DeepSeek-v3. The introspective and inward-focused cognitive stance associated with LE likely fosters more deliberate and fine-grained judgment, enhancing performance in multi-level grading tasks. High Neuroticism (HN) surpasses the default for both GPT-4o and Llama-3-70B, suggesting that the vigilance and heightened sensitivity to detail the characteristic of HN help some LLMs establish more precise decision boundaries.
Low Openness (LO) and Low Neuroticism (LN) yield moderate gains, outperforming the baseline on the three datasets in certain models (e.g., DeepSeek-v3 for both LO and LN, GPT-4o for LN)
. Additionally, the performance of LN appears to be influenced by CoT: under the with-CoT setting, LN consistently outperforms the default condition. High Extraversion (HE) is generally weak, yet on Llama-3-8B it has a strong performance under the with-CoT instruction. Low Conscientiousness (LC) also performs weakly overall, yet unexpectedly shows strong results on DeepSeek-v3.

\textbf{Key findings.} Low Agreeableness consistently aligns with human judgement better against the default setting, while traits such as High Conscientiousness, Low Extraversion, and High Neuroticism also enhance alignment in model-specific ways. 

\subsection{RQ2: Confidence Reliability}
\begin{figure}[htbp]
    \centering

    \begin{subfigure}{\textwidth}
        \centering
        \includegraphics[width=0.78\textwidth]{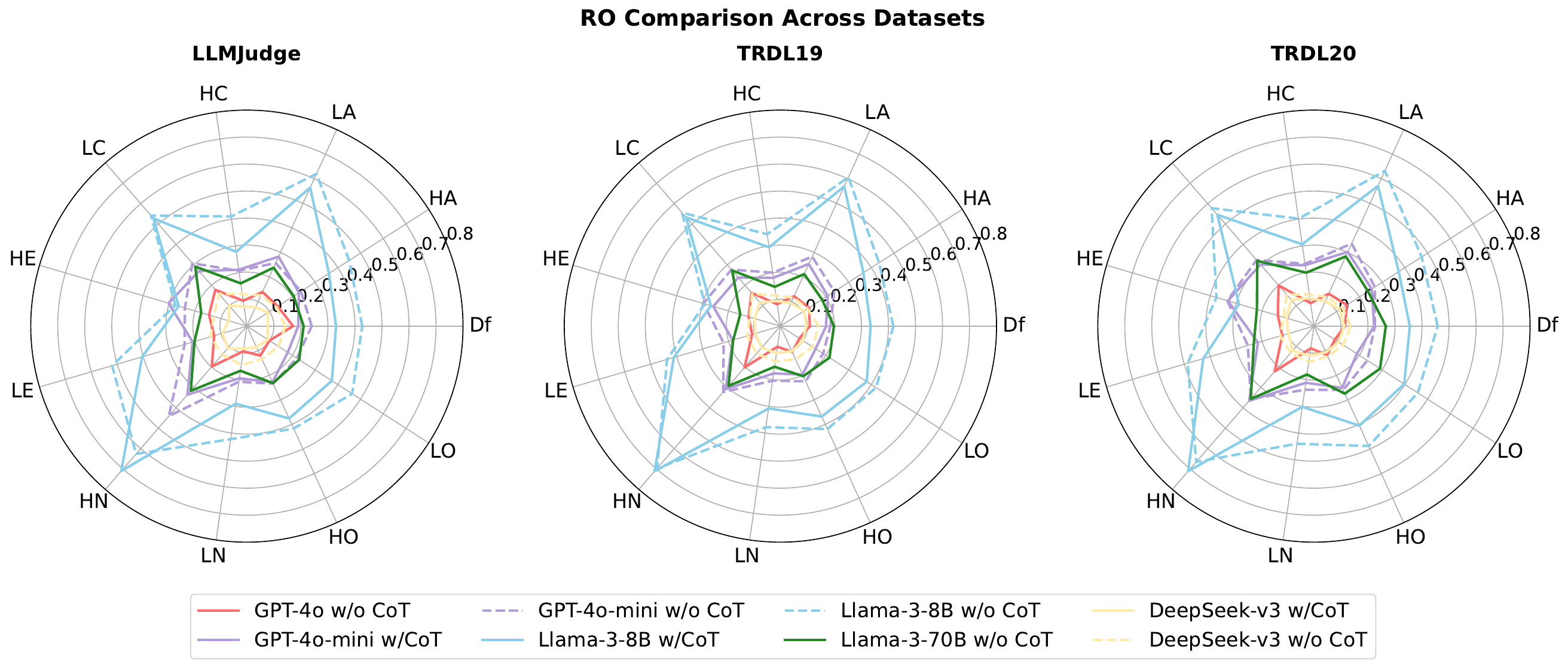}
    \end{subfigure}

    \vspace{1cm}

    \begin{subfigure}{\textwidth}
        \centering
        \includegraphics[width=0.78\textwidth]{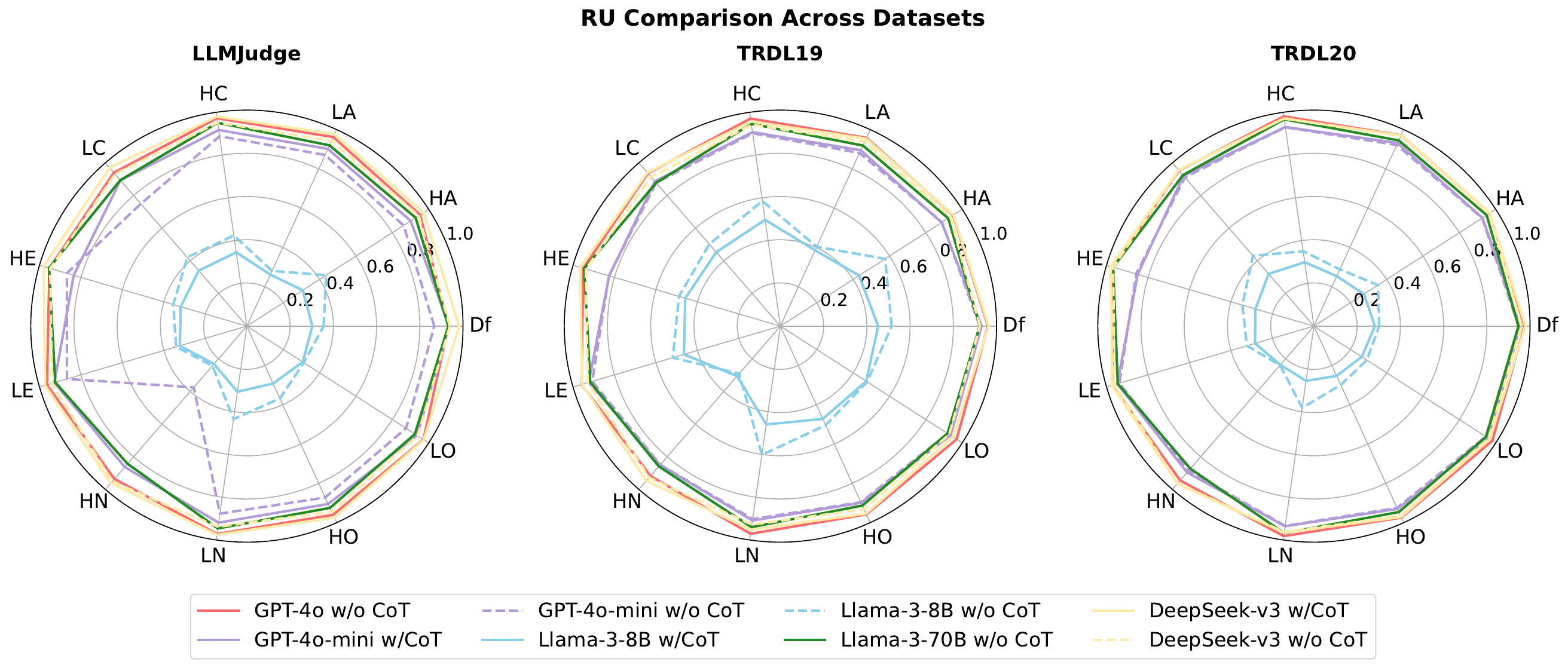}
    \end{subfigure}

    \vspace{1cm}

    \begin{subfigure}{\textwidth}
        \centering
        \includegraphics[width=0.78\textwidth]{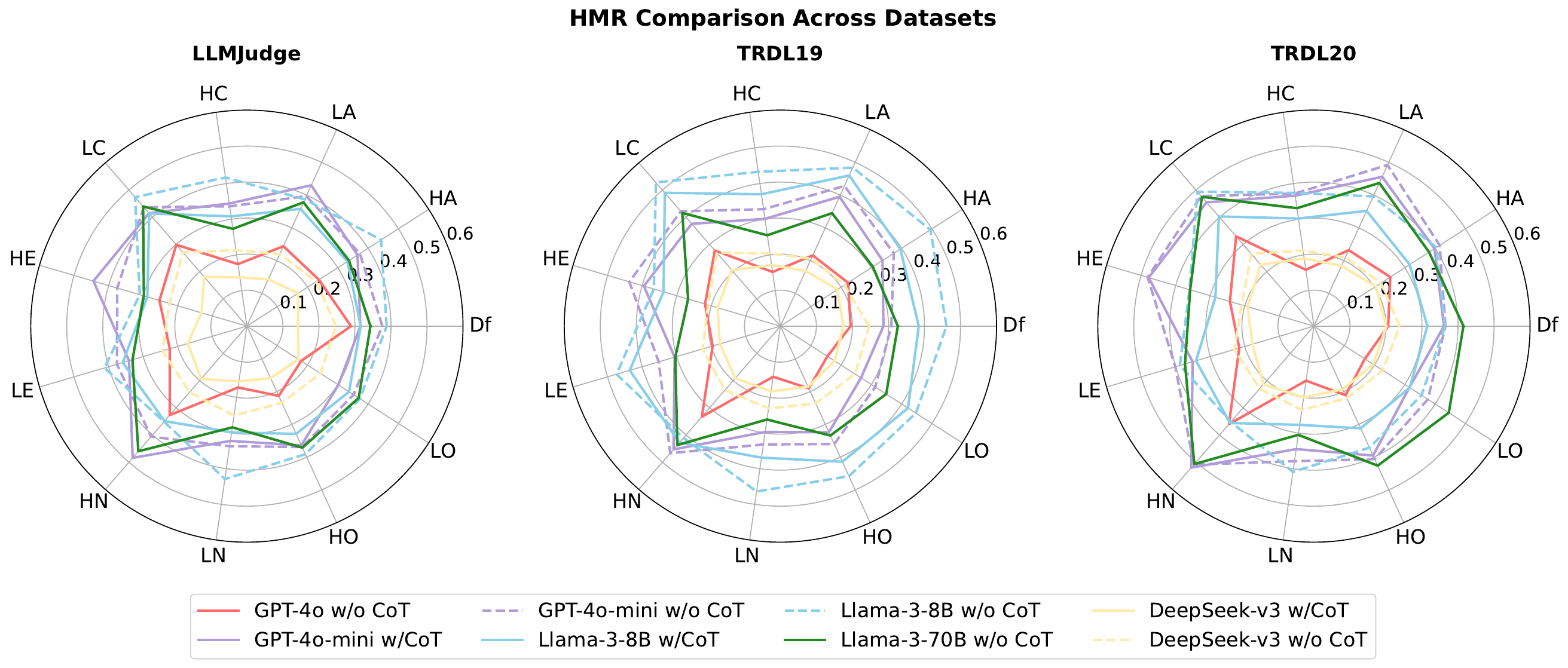}
    \end{subfigure}

    \caption{Comparison of RO, RU, and HMR metrics across different personality dimensions and model configurations. Each radar chart shows performance across 11 personality configurations (Df: Default, HA/LA: High/Low Agreeableness, HC/LC: High/Low Conscientiousness, HE/LE: High/Low Extraversion, HN/LN: High/Low Neuroticism, HO/LO: High/Low Openness) for three datasets (LLMJudge, TRDL19, TRDL20). RO measures rank order correlation, RU measures rank utility, and HMR measures human-machine relevance agreement.}
    \label{fig:conf_res}
\end{figure}

Figure~\ref{fig:conf_res} presents the comparison of evaluation metrics  (RO, RU, HMR)  across different personality dimensions and model configurations. From Figure~\ref{fig:conf_res} one can observe the follows. 

From a model level perspective, a comparative examination of RO, RU, and HMR across multiple LLMs reveals substantial heterogeneity in metacognitive self calibration. Taken together, RO and RU suggest that most models, particularly those with stronger language capabilities and larger parameter scales, exhibit a systematic tendency toward overconfidence. These models tend to assign high confidence to their predictions even when they are incorrect, leading to an imbalanced calibration profile in which strong confidence assertion is not matched by adequate error recognition. As a consequence, their overall calibration, as captured by HMR, remains suboptimal despite their superior linguistic and reasoning performance. In contrast, smaller models such as Llama-3-8B display a markedly different calibration pattern. They appear more effective at suppressing overconfidence, as reflected by a greater willingness to acknowledge uncertainty or potential error. However, this advantage comes at a pronounced cost in terms of underconfidence control. Specifically, these models often fail to assert sufficient confidence even when their judgments are correct, suggesting a generally lower baseline confidence across their output distributions. As a result, their improved error awareness does not translate into a superior overall balance between overconfidence and underconfidence, leaving their HMR comparable to, rather than better than, that of larger models that maintain a more moderate calibration profile.

From a personality-oriented perspective, the analysis of RO, RU, and HMR reveals distinct and asymmetric patterns of performance. Low Conscientiousness achieves superiority over the default baseline in all 24 instances for both RO and HMR, making it the only personality trait to demonstrate global perfection across two metrics. This suggests that inducing a less rigorous and accountable cognitive stance in LLMs may help suppress their overconfidence bias. Though at the expense of weaker confidence assertion, as reflected in RU, the gain from reducing overconfidence outweighs the underconfidence cost, yielding improved overall calibration as measured by HMR. A possible explain is that: when induced with LC-related traits such as cognitive flexibility, reduced discipline, and limited attention to detail (refer to Table~\ref{tab:big5-prompts} in the appendix), the model becomes less compelled to uphold the authority of its own judgments, thereby mitigating the overconfidence bias. 

High Neuroticism also effectively suppresses overconfidence and achieves balanced control between overconfidence and underconfidence, for all LLMs except DeepSeek-v3. HN related traits, such as anxiety and self-doubt, induce a state of heightened vigilance and self-monitoring, effectively suppressing overconfidence and outperforming the default setting in most cases. However, this cautious stance reduces confidence in correct judgments, leading to weaker RU performance, which is consistent with the observations reported by~\citet{jacobs2012}. Despite this, HN achieves superior overall calibration (HMR) through stronger control of overconfidence. By contrast, High Conscientiousness shows the best RU performance across models such as GPT-4o, Llama-3-70B, and Llama-3-8B. This suggests that traits linked to diligence, control, and self-discipline encourage \textit{assertiveness}~\citep{Reed2020SelfRegulation, EvansFrankish2009} in correct decisions and consistent confidence reinforcement. However, this self-assuredness limits adaptability, resulting in poorer RO and HMR outcomes. 

The performance of other personality conditions depends on specific models or fluctuates across datasets. For example, High Extraversion has been previously reported to be linked with overconfidence~\citep{wolfe1990, anderson2014,schaefer2004,soh2013}, but in our results, except for those on Llama-3-70B, we did not observe a clear disadvantage of HE in suppressing overconfidence. The diversity of results across models underscores that personality conditioning interacts nonlinearly with model scale and architecture, producing different metacognitive trade-offs between confidence suppression and assertion.

\textbf{Key findings.} Compared to the default setting, traits such as Low Conscientiousness and High Neuroticism effectively suppress overconfidence and improve overall balance (HMR), whereas High Conscientiousness enhances confidence assertion but at the cost of reduced supression against overconfidence. 

\subsection{RQ3: Personality Features for Label Classification}

\begin{table*}[htbp]
    \centering
    \caption{Performance comparison across ML models, conditions, and metrics. Values show mean (SD). Oracle denotes the highest score achieved by any single personality condition induced by the LLM for a given metric under the current setting. Cell colors indicate the corresponding personality conditions.}
    \label{tab:comparison}
    \resizebox{\textwidth}{!}{%
    \begin{tabular}{|l|l|l|cc|cc|cc|cc|cc|cc|}
    \hline
    \multirow{2}{*}{\textbf{LLM}} & \multirow{2}{*}{\textbf{CoT}} & \multirow{2}{*}{\textbf{Metric}} & \multicolumn{2}{c|}{\textbf{RF}} & \multicolumn{2}{c|}{\textbf{RF w/o conf.}} & \multicolumn{2}{c|}{\textbf{XGB}} & \multicolumn{2}{c|}{\textbf{LGBM}} & \multicolumn{2}{c|}{\textbf{Ord. Log.}} & \multicolumn{2}{c|}{\textbf{Oracle}} \\
    \cline{4-15}
     & & & Mean & SD & Mean & SD & Mean & SD & Mean & SD & Mean & SD & Mean & SD \\
    \hline
    \multirow{6}{*}{\textbf{GPT-4o-mini}} & \multirow{3}{*}{w/ CoT} & $\kappa$ & \textbf{0.371} & \textbf{(0.011)} & \textbf{0.347} & \textbf{(0.010)} & 0.326 & (0.013) & 0.322 & (0.015) & 0.287 & (0.013) & \cellcolor{milktea!30}0.343 & \cellcolor{milktea!30}(0.003) \\
     & & QWK & \textbf{0.626} & \textbf{(0.008)} & 0.601 & (0.007) & 0.572 & (0.019) & 0.569 & (0.018) & 0.565 & (0.014) & \cellcolor{royalblue!30}0.604 & \cellcolor{royalblue!30}(0.003) \\
     & & F1 & \textbf{0.514} & \textbf{(0.009)} & 0.497 & (0.014) & 0.476 & (0.011) & 0.472 & (0.012) & 0.391 & (0.011) & \cellcolor{milktea!30}0.498 & \cellcolor{milktea!30}(0.003) \\
    \cline{2-15}
     & \multirow{3}{*}{w/o CoT} & $\kappa$ & \textbf{0.364} & \textbf{(0.013)} & \textbf{0.321} & \textbf{(0.015)} & \textbf{0.319} & \textbf{(0.015)} & \textbf{0.318} & \textbf{(0.016)} & 0.277 & (0.010) & \cellcolor{royalblue!30}0.311 & \cellcolor{royalblue!30}(0.004) \\
     & & QWK & \textbf{0.622} & \textbf{(0.012)} & \textbf{0.582} & \textbf{(0.019)} & 0.565 & (0.021) & 0.566 & (0.020) & \textbf{0.579} & \textbf{(0.011)} & \cellcolor{royalblue!30}0.573 & \cellcolor{royalblue!30}(0.003) \\
     & & F1 & \textbf{0.508} & \textbf{(0.011)} & 0.469 & (0.011) & 0.473 & (0.012) & 0.471 & (0.013) & 0.383 & (0.009) & \cellcolor{royalblue!30}0.477 & \cellcolor{royalblue!30}(0.003) \\
    \hline
    \multirow{6}{*}{\textbf{LLama-3-8B}} & \multirow{3}{*}{w/ CoT} & $\kappa$ & \textbf{0.262} & \textbf{(0.016)} & \textbf{0.246} & \textbf{(0.017)} & \textbf{0.261} & \textbf{(0.017)} & \textbf{0.259} & \textbf{(0.018)} & \textbf{0.244} & \textbf{(0.012)} & \cellcolor{royalblue!30}0.114 & \cellcolor{royalblue!30}(0.003) \\
     & & QWK & \textbf{0.494} & \textbf{(0.018)} & \textbf{0.471} & \textbf{(0.022)} & \textbf{0.464} & \textbf{(0.021)} & \textbf{0.463} & \textbf{(0.025)} & \textbf{0.499} & \textbf{(0.016)} & \cellcolor{royalblue!30}0.385 & \cellcolor{royalblue!30}(0.003) \\
     & & F1 & \textbf{0.417} & \textbf{(0.015)} & \textbf{0.388} & \textbf{(0.016)} & \textbf{0.429} & \textbf{(0.014)} & \textbf{0.427} & \textbf{(0.015)} & \textbf{0.378} & \textbf{(0.011)} & \cellcolor{royalblue!30}0.273 & \cellcolor{royalblue!30}(0.002) \\
    \cline{2-15}
     & \multirow{3}{*}{w/o CoT} & $\kappa$ & \textbf{0.267} & \textbf{(0.016)} & \textbf{0.224} & \textbf{(0.021)} & \textbf{0.259} & \textbf{(0.015)} & \textbf{0.260} & \textbf{(0.016)} & \textbf{0.243} & \textbf{(0.012)} & \cellcolor{royalblue!30}0.113 & \cellcolor{royalblue!30}(0.002) \\
     & & QWK & \textbf{0.478} & \textbf{(0.020)} & \textbf{0.387} & \textbf{(0.031)} & \textbf{0.433} & \textbf{(0.023)} & \textbf{0.437} & \textbf{(0.023)} & \textbf{0.495} & \textbf{(0.011)} & \cellcolor{royalblue!30}0.340 & \cellcolor{royalblue!30}(0.003) \\
     & & F1 & \textbf{0.431} & \textbf{(0.015)} & \textbf{0.379} & \textbf{(0.020)} & \textbf{0.428} & \textbf{(0.013)} & \textbf{0.428} & \textbf{(0.015)} & \textbf{0.369} & \textbf{(0.008)} & \cellcolor{myred!20}0.251 & \cellcolor{myred!20}(0.002) \\
    \hline
    \multirow{6}{*}{\textbf{Deepseek-v3}} & \multirow{3}{*}{w/ CoT} & $\kappa$ & \textbf{0.372} & \textbf{(0.012)} & \textbf{0.364} & \textbf{(0.012)} & 0.334 & (0.022) & 0.333 & (0.022) & 0.320 & (0.011) & \cellcolor{myyellow!30}0.354 & \cellcolor{myyellow!30}(0.004) \\
     & & QWK & 0.623 & (0.011) & 0.613 & (0.011) & 0.582 & (0.020) & 0.582 & (0.022) & 0.598 & (0.011) & \cellcolor{royalblue!30}0.631 & \cellcolor{royalblue!30}(0.003) \\
     & & F1 & \textbf{0.515} & \textbf{(0.014)} & \textbf{0.506} & \textbf{(0.014)} & 0.480 & (0.028) & 0.481 & (0.027) & 0.409 & (0.007) & \cellcolor{mikan!15}0.503 & \cellcolor{mikan!15}(0.003) \\
    \cline{2-15}
     & \multirow{3}{*}{w/o CoT} & $\kappa$ & \textbf{0.376} & \textbf{(0.009)} & \textbf{0.363} & \textbf{(0.009)} & 0.338 & (0.016) & 0.333 & (0.016) & 0.307 & (0.009) & \cellcolor{senzaimidori!20}0.351 & \cellcolor{senzaimidori!20}(0.004) \\
     & & QWK & \textbf{0.623} & \textbf{(0.009)} & \textbf{0.614} & \textbf{(0.011)} & 0.591 & (0.021) & 0.589 & (0.021) & \textbf{0.607} & \textbf{(0.009)} & \cellcolor{senzaimidori!20}0.604 & \cellcolor{senzaimidori!20}(0.003) \\
     & & F1 & \textbf{0.515} & \textbf{(0.009)} & \textbf{0.503} & \textbf{(0.008)} & 0.469 & (0.030) & 0.466 & (0.029) & 0.398 & (0.006) & \cellcolor{senzaimidori!20}0.498 & \cellcolor{senzaimidori!20}(0.003) \\
    \hline
    \end{tabular}
    }
    \end{table*}

Table~\ref{tab:comparison} presents the results of Experiment 2, where Ord. Log. denotes the classifier based on ordinal logistic regression, and Oracle denotes the highest score achieved by any single personality condition induced by the LLM under the current setting for a given metric. For example, under the with-CoT setting, the Oracle scores of Deepseek-v3 on $\kappa$, QWK, and F1 are derived from the HC, LA, and LN personality conditions, respectively, with each achieving the highest performance among the eleven personality variants for the corresponding metric; in contrast, under the without-CoT setting, the Oracle scores of GPT-4o-mini across all three metrics are consistently attributed to the LA condition, which obtains the highest score among all eleven variants for each metric.

Based on Table~\ref{tab:comparison}, we have the following findings. Across learning algorithms that consume the full personality-informed feature set, Random Forest delivers the most robust performance on $\kappa$, QWK, and F1, with tight dispersion across trials. Ordinal Logistic is weak on $\kappa$ and F1, under the CoT setting, however it surpasses the Oracle baseline on QWK. XGB and LGBM are generally the weakest performers. Apart from isolated cases, their results on models other than Llama 3 8B fall below the Oracle baseline, whereas on Llama 3 8B they show clear improvements over the Oracle baseline.

Taking Random Forest as the representative case, this finding indicates that aggregating relevance scores and confidence estimates from all personality conditions as features yields systematic gains with respect to human judgement alignment over the Oracle baseline, i.e., the best single personality per metric. These gains are achieved with only limited human labeled data, and they are most pronounced for Llama 3 8B, where cross personality integration compensates for weaker base judgments by exploiting complementary cues not available from any single personality condition. The same pattern holds under both with CoT and without CoT settings, although the relative magnitude of improvement is model dependent.

To evaluate the predictive utility of the confidence scores produced by each personality condition, we conducted an ablation study. In Table~\ref{tab:comparison}, RF w/o conf. denotes the variant that trains the Random Forest using only the predicted relevance scores from the eleven personality conditions as features, excluding confidence. The ablation study confirms that confidence carries independent predictive value beyond scores. Removing confidence from the Random Forest features yields consistent declines on $\kappa$, QWK, and F1 across models and CoT settings, indicating that self reported confidence is not a redundant proxy for scores but an informative signal that improves class separability in the supervised stage.

\textbf{Key findings.} Personality-conditioned prediction scores and confidence values are effective features for relevance classification when paired with suitable machine learning algorithms. Aggregating outputs from all personality conditions improves performance over the Oracle baseline, even with limited labeled data, especially for weaker models like Llama-3-8B. Ablation results show that confidence adds predictive value beyond relevance scores.


\section{Discussion and Conclusion}
\subsection{Discussion and Open Questions}

In this study, we investigated the behavior of LLMs simulating different personality traits in the context of relevance assessment, focusing specifically on the alignment of their decisions with human preferences and the reliability of their self-reported confidence. In Experiment 1, we identified several cross-model patterns. Notably, assessors induced with low Agreeableness consistently achieved higher alignment with human judgments in terms of $\kappa$, QWK, and F1 compared to the default setting. However, the underlying mechanisms behind this phenomenon remain an open question. For instance, it is unclear whether specific tokens in the low-Agreeableness prompt elicited more critical reasoning behaviors in the model, or whether the decision styles of most human annotators inherently resemble those of a low-Agreeableness profile. Another consistent observation from Experiment 1 is that simulated personalities with Low Conscientiousness and High Neuroticism effectively suppress overconfidence while maintaining a relatively balanced calibration between overconfidence and underconfidence. However, our current design only asked LLMs to report confidence in their own decisions. Future work may explore whether these two personality conditions can reliably estimate confidence in the decisions made by other personalities, including the default setting, and whether they continue to preserve balanced calibration in this broader context. If so, this would suggest a promising and psychologically grounded pathway for improving LLM confidence calibration in a more generalizable manner.

\subsection{Conclusion}

In this study, we presented a novel, psychologically grounded framework for relevance evaluation, shifting the paradigm from generic prompting to personality-conditioned cognitive modeling. By systematically infusing Big Five personality traits into large language models, we explored how distinct cognitive profiles influence both the accuracy of relevance judgments and the reliability of confidence calibration. We conducted experiments spanning five diverse LLMs and three IR test collections (TREC DL 2019, 2020, and LLMJudge).


Synthesizing our empirical findings, we identify a distinct trade-off between judgmental alignment and calibration reliability. Specifically, while the cognitive profile associated with Low Agreeableness demonstrated superior efficacy in replicating expert human relevance judgments, traits such as Low Conscientiousness and High Neuroticism were necessary to effectively mitigate systematic overconfidence. We attribute these improvements to the distinct decision-making thresholds established by personality conditioning: the strict evaluation criteria associated with Low Agreeableness enhance the model's discriminative precision, while the conservative estimation tendencies linked to High Neuroticism counteract the model's inherent bias toward unwarranted certainty. Notably, this intervention is particularly critical for high-capability models (e.g., GPT-4o), which exhibited a stronger propensity for overconfidence despite their advanced reasoning capabilities.

This dynamics suggests that no single personality condition is universally optimal across all evaluation metrics; rather, developing high-quality automated evaluators requires balancing discriminative acuity with robust metacognitive calibration. Additionally, our work demonstrates that this cognitive diversity is not noise, but a valuable signal. By aggregating predictions across simulated personalities, we showed that personality-derived confidence features provide unique, complementary information that significantly enhances relevance classification, outperforming single-persona baselines even in low-resource scenarios.

Broadly, this work contributes to the emerging field of Machine Psychology~\citep{hagendorff2023machine} by providing concrete evidence that abstract psychological constructs can be operationalized to modulate LLM behavior in predictable ways. Within the Information Retrieval community, our findings offer a practical pathway toward trustworthy evaluation systems that transcend mere prediction accuracy, evolving instead into agents capable of explicitly signaling when and why they might be incorrect. Looking ahead, future work will extend this framework to investigate how personality interactions—such as debates between diverse agents—might further refine the boundaries of automated decision-making.


\bibliographystyle{ACM-Reference-Format}
\bibliography{sample-base}

\appendix
\section{The Confidence Reporting Instruction Used in the Study}
\label{app:conf_inst}
\begin{table}[htbp]
\centering
\caption{The confidence reporting instruction used in this study. For both the with-CoT and without-CoT versions, we used the following instruction to elicit confidence scores.}
\begin{myframe}
\begin{quote}
\noindent\emph{ You are an expert in evaluating the relevance of text passages to user queries. }
\noindent\emph{Your task is to assign a relevance score to a passage based on how well it addresses the information need expressed in a query. }
\noindent\emph{Use the following scale:}

\noindent\emph{[3] Perfectly relevant: The passage is fully focused on the query and provides a clear and complete answer.}

\noindent\emph{[2] Highly relevant: The passage provides some relevant information but may include extraneous details or lack clarity.}

\noindent\emph{[1] Related: The passage is tangentially related to the query but does not answer it.}

\noindent\emph{[0] Irrelevant: The passage has no connection to the query.}

\noindent\emph{Query: \textcolor{blue}{\texttt{\{query\}}}}
\noindent\emph{Passage: \textcolor{blue}{\texttt{\{document\}}}}

\noindent\emph{Your given relevance score is \textcolor{blue}{\{pblueicted\_score\}}, please give a confidence score for this answer representing how confident you believe this answer is correct. 0 means you have no confidence at all, and 100 means you have absolute confidence. ONLY return a number from 0 to 100 to show your confidence to your answer and do not return any other content.}

\end{quote}
\end{myframe}
\label{tab:conf_prompt}
\end{table}
Table~\ref{tab:conf_prompt} presents the confidence reporting instruction used in this study. For both the with-CoT and without-CoT versions, we used the following instruction to elicit confidence scores.

\section{The full results of Human Alignment and Confidence Reliability.}
\label{app:full_res}
\begin{table*}[htbp]
    \scriptsize
    \centering
    \caption{Performance on human alignment of five LLMs under w/CoT and w/o CoT across datasets ($\kappa$, QWK, F1). Cells are shaded gray for the default condition and highlighted with color if performance exceeds the default.}
    \label{tab:alignment_res}
    \begin{tabular}{c|c|ccc|ccc|ccc|ccc|ccc|ccc}
    \hline
    \multirow{3}{*}{\textbf{Model}} & \multirow{3}{*}{\textbf{P.}}
    & \multicolumn{6}{c|}{\textbf{LLMJudge}}
    & \multicolumn{6}{c|}{\textbf{TRDL19}}
    & \multicolumn{6}{c}{\textbf{TRDL20}} \\
    \cline{3-20}
     & & \multicolumn{3}{c|}{\textbf{w/CoT}} & \multicolumn{3}{c|}{\textbf{w/o CoT}}
     & \multicolumn{3}{c|}{\textbf{w/CoT}} & \multicolumn{3}{c|}{\textbf{w/o CoT}}
     & \multicolumn{3}{c|}{\textbf{w/CoT}} & \multicolumn{3}{c}{\textbf{w/o CoT}} \\
    \cline{3-20}
     &
     & \textbf{$\kappa$} & \textbf{QWK} & \textbf{F1} & \textbf{$\kappa$} & \textbf{QWK} & \textbf{F1}
     & \textbf{$\kappa$} & \textbf{QWK} & \textbf{F1} & \textbf{$\kappa$} & \textbf{QWK} & \textbf{F1}
     & \textbf{$\kappa$} & \textbf{QWK} & \textbf{F1} & \textbf{$\kappa$} & \textbf{QWK} & \textbf{F1} \\
    \hline
    \multirow{11}{*}{G.}
     & \cellcolor{gray!15}Df. & & & & \cellcolor{gray!15}0.306 & \cellcolor{gray!15}0.562 & \cellcolor{gray!15}0.423 & & & & \cellcolor{gray!15}0.253 & \cellcolor{gray!15}0.463 & \cellcolor{gray!15}0.377 & & & & \cellcolor{gray!15}0.364 & \cellcolor{gray!15}0.534 & \cellcolor{gray!15}0.458 \\
     & \cellcolor{sakurapink!15}HA & & & & 0.266 & 0.541 & \cellcolor{sakurapink!15}0.427 & & & & 0.218 & 0.462 & 0.362 & & & & 0.334 & 0.534 & 0.450 \\
     & \cellcolor{royalblue!30}LA & & & & \cellcolor{royalblue!30}0.324 & \cellcolor{royalblue!30}0.564 & \cellcolor{royalblue!30}0.428 & & & & \cellcolor{royalblue!30}0.291 & \cellcolor{royalblue!30}0.467 & \cellcolor{royalblue!30}0.407 & & & & \cellcolor{royalblue!30}0.391 & \cellcolor{royalblue!30}0.544 & \cellcolor{royalblue!30}0.467 \\
     & \cellcolor{myyellow!30}HC & & & & \cellcolor{myyellow!30}0.325 & \cellcolor{myyellow!30}0.581 & \cellcolor{myyellow!30}0.444 & & & & \cellcolor{myyellow!30}0.266 & \cellcolor{myyellow!30}0.475 & \cellcolor{myyellow!30}0.384 & & & & \cellcolor{myyellow!30}0.382 & \cellcolor{myyellow!30}0.539 & \cellcolor{myyellow!30}0.466 \\
     & \cellcolor{myred!20}LC & & & & 0.297 & \cellcolor{myred!20}0.563 & \cellcolor{myred!20}0.433 & & & & \cellcolor{myred!20}0.260 & \cellcolor{myred!20}0.476 & \cellcolor{myred!20}0.391 & & & & 0.349 & 0.528 & 0.457 \\
     & \cellcolor{emraldgreen!15}HE & & & & 0.266 & 0.538 & 0.419 & & & & 0.240 & \cellcolor{emraldgreen!15}0.478 & 0.371 & & & & \cellcolor{emraldgreen!15}0.370 & \cellcolor{emraldgreen!15}0.541 & \cellcolor{emraldgreen!15}0.467 \\
     & \cellcolor{milktea!30}LE & & & & \cellcolor{milktea!30}0.317 & \cellcolor{milktea!30}0.572 & \cellcolor{milktea!30}0.435 & & & & \cellcolor{milktea!30}0.276 & \cellcolor{milktea!30}0.475 & \cellcolor{milktea!30}0.393 & & & & \cellcolor{milktea!30}0.375 & \cellcolor{milktea!30}0.553 & 0.458 \\
     & \cellcolor{senzaimidori!20}HN & & & & \cellcolor{senzaimidori!20}0.318 & \cellcolor{senzaimidori!20}0.576 & \cellcolor{senzaimidori!20}0.430 & & & & \cellcolor{senzaimidori!20}0.270 & \cellcolor{senzaimidori!20}0.477 & \cellcolor{senzaimidori!20}0.393 & & & & \cellcolor{senzaimidori!20}0.377 & \cellcolor{senzaimidori!20}0.540 & \cellcolor{senzaimidori!20}0.463 \\
     & \cellcolor{mikan!15}LN & & & & 0.304 & 0.561 & \cellcolor{mikan!15}0.434 & & & & \cellcolor{mikan!15}0.262 & \cellcolor{mikan!15}0.483 & \cellcolor{mikan!15}0.384 & & & & \cellcolor{mikan!15}0.376 & \cellcolor{mikan!15}0.538 & \cellcolor{mikan!15}0.469 \\
     & \cellcolor{myviolet!15}HO & & & & 0.270 & 0.546 & 0.420 & & & & 0.231 & 0.461 & 0.365 & & & & 0.341 & 0.525 & 0.452 \\
     & \cellcolor{mylightblue!15}LO & & & & \cellcolor{mylightblue!15}0.331 & \cellcolor{mylightblue!15}0.575 & \cellcolor{mylightblue!15}0.440 & & & & \cellcolor{mylightblue!15}0.267 & \cellcolor{mylightblue!15}0.472 & \cellcolor{mylightblue!15}0.379 & & & & 0.352 & 0.534 & 0.444 \\
    \hline
    \multirow{11}{*}{G.m}
     & \cellcolor{gray!15}Df. & \cellcolor{gray!15}0.264 & \cellcolor{gray!15}0.524 & \cellcolor{gray!15}0.422 & \cellcolor{gray!15}0.262 & \cellcolor{gray!15}0.524 & \cellcolor{gray!15}0.423 & \cellcolor{gray!15}0.246 & \cellcolor{gray!15}0.467 & \cellcolor{gray!15}0.383 & \cellcolor{gray!15}0.236 & \cellcolor{gray!15}0.447 & \cellcolor{gray!15}0.389 & \cellcolor{gray!15}0.310 & \cellcolor{gray!15}0.517 & \cellcolor{gray!15}0.432 & \cellcolor{gray!15}0.294 & \cellcolor{gray!15}0.489 & \cellcolor{gray!15}0.418 \\
     & \cellcolor{sakurapink!15}HA & 0.260 & 0.522 & \cellcolor{sakurapink!15}0.424 & 0.238 & 0.515 & 0.410 & \cellcolor{sakurapink!15}0.249 & \cellcolor{sakurapink!15}0.473 & \cellcolor{sakurapink!15}0.391 & 0.213 & 0.441 & 0.380 & 0.304 & \cellcolor{sakurapink!15}0.520 & 0.426 & 0.277 & 0.485 & 0.414 \\
     & \cellcolor{royalblue!30}LA & \cellcolor{royalblue!30}0.283 & \cellcolor{royalblue!30}0.534 & \cellcolor{royalblue!30}0.433 & \cellcolor{royalblue!30}0.284 & \cellcolor{royalblue!30}0.535 & \cellcolor{royalblue!30}0.436 & \cellcolor{royalblue!30}0.283 & \cellcolor{royalblue!30}0.474 & \cellcolor{royalblue!30}0.412 & \cellcolor{royalblue!30}0.264 & \cellcolor{royalblue!30}0.460 & \cellcolor{royalblue!30}0.411 & \cellcolor{royalblue!30}0.324 & \cellcolor{royalblue!30}0.527 & 0.432 & \cellcolor{royalblue!30}0.306 & \cellcolor{royalblue!30}0.491 & \cellcolor{royalblue!30}0.427 \\
     & \cellcolor{myyellow!30}HC & 0.258 & 0.520 & \cellcolor{myyellow!30}0.426 & \cellcolor{myyellow!30}0.266 & \cellcolor{myyellow!30}0.528 & 0.423 & \cellcolor{myyellow!30}0.260 & \cellcolor{myyellow!30}0.471 & \cellcolor{myyellow!30}0.398 & 0.236 & \cellcolor{myyellow!30}0.452 & \cellcolor{myyellow!30}0.393 & \cellcolor{myyellow!30}0.312 & \cellcolor{myyellow!30}0.527 & 0.429 & 0.293 & 0.484 & 0.417 \\
     & \cellcolor{myred!20}LC & 0.249 & 0.509 & 0.417 & 0.250 & 0.511 & 0.418 & 0.235 & 0.464 & 0.380 & 0.214 & 0.432 & 0.382 & 0.292 & \cellcolor{myred!20}0.518 & 0.421 & 0.268 & 0.468 & 0.409 \\
     & \cellcolor{emraldgreen!15}HE & 0.251 & 0.514 & 0.420 & 0.248 & 0.520 & 0.414 & 0.223 & 0.458 & 0.375 & 0.206 & 0.436 & 0.374 & 0.291 & 0.511 & 0.421 & 0.268 & 0.478 & 0.408 \\
     & \cellcolor{milktea!30}LE & \cellcolor{milktea!30}0.293 & \cellcolor{milktea!30}0.544 & \cellcolor{milktea!30}0.443 & \cellcolor{milktea!30}0.272 & \cellcolor{milktea!30}0.531 & \cellcolor{milktea!30}0.433 & \cellcolor{milktea!30}0.270 & \cellcolor{milktea!30}0.479 & \cellcolor{milktea!30}0.400 & \cellcolor{milktea!30}0.238 & \cellcolor{milktea!30}0.452 & \cellcolor{milktea!30}0.393 & \cellcolor{milktea!30}0.324 & \cellcolor{milktea!30}0.525 & \cellcolor{milktea!30}0.437 & \cellcolor{milktea!30}0.298 & \cellcolor{milktea!30}0.491 & \cellcolor{milktea!30}0.422 \\
     & \cellcolor{senzaimidori!20}HN & 0.255 & 0.514 & 0.420 & \cellcolor{senzaimidori!20}0.269 & \cellcolor{senzaimidori!20}0.526 & \cellcolor{senzaimidori!20}0.429 & 0.230 & 0.457 & 0.381 & \cellcolor{senzaimidori!20}0.247 & \cellcolor{senzaimidori!20}0.448 & \cellcolor{senzaimidori!20}0.397 & 0.299 & \cellcolor{senzaimidori!20}0.523 & 0.427 & \cellcolor{senzaimidori!20}0.304 & \cellcolor{senzaimidori!20}0.494 & \cellcolor{senzaimidori!20}0.422 \\
     & \cellcolor{mikan!15}LN & \cellcolor{mikan!15}0.272 & \cellcolor{mikan!15}0.527 & \cellcolor{mikan!15}0.429 & 0.242 & 0.515 & 0.407 & \cellcolor{mikan!15}0.255 & \cellcolor{mikan!15}0.479 & \cellcolor{mikan!15}0.391 & 0.219 & \cellcolor{mikan!15}0.449 & 0.378 & \cellcolor{mikan!15}0.316 & \cellcolor{mikan!15}0.528 & 0.432 & 0.287 & \cellcolor{mikan!15}0.491 & 0.415 \\
     & \cellcolor{myviolet!15}HO & 0.252 & 0.511 & \cellcolor{myviolet!15}0.423 & 0.230 & 0.511 & 0.402 & 0.209 & 0.445 & 0.365 & 0.186 & 0.424 & 0.358 & 0.288 & 0.517 & 0.422 & 0.269 & 0.482 & 0.409 \\
     & \cellcolor{mylightblue!15}LO & \cellcolor{mylightblue!15}0.283 & \cellcolor{mylightblue!15}0.534 & \cellcolor{mylightblue!15}0.435 & \cellcolor{mylightblue!15}0.268 & \cellcolor{mylightblue!15}0.530 & 0.420 & \cellcolor{mylightblue!15}0.253 & \cellcolor{mylightblue!15}0.472 & 0.382 & 0.225 & \cellcolor{mylightblue!15}0.452 & 0.378 & \cellcolor{mylightblue!15}0.323 & \cellcolor{mylightblue!15}0.525 & \cellcolor{mylightblue!15}0.436 & 0.290 & \cellcolor{mylightblue!15}0.491 & \cellcolor{mylightblue!15}0.419 \\
    \hline
    \multirow{11}{*}{L.8b}
     & \cellcolor{gray!15}Df. & \cellcolor{gray!15}0.163 & \cellcolor{gray!15}0.322 & \cellcolor{gray!15}0.288 & \cellcolor{gray!15}0.088 & \cellcolor{gray!15}0.281 & \cellcolor{gray!15}0.231 & \cellcolor{gray!15}0.111 & \cellcolor{gray!15}0.204 & \cellcolor{gray!15}0.262 & \cellcolor{gray!15}0.095 & \cellcolor{gray!15}0.206 & \cellcolor{gray!15}0.248 & \cellcolor{gray!15}0.134 & \cellcolor{gray!15}0.255 & \cellcolor{gray!15}0.259 & \cellcolor{gray!15}0.064 & \cellcolor{gray!15}0.258 & \cellcolor{gray!15}0.201 \\
     & \cellcolor{sakurapink!15}HA & \cellcolor{sakurapink!15}0.200 & \cellcolor{sakurapink!15}0.405 & \cellcolor{sakurapink!15}0.343 & 0.040 & 0.237 & 0.175 & \cellcolor{sakurapink!15}0.145 & \cellcolor{sakurapink!15}0.279 & \cellcolor{sakurapink!15}0.296 & 0.070 & 0.186 & 0.204 & \cellcolor{sakurapink!15}0.162 & \cellcolor{sakurapink!15}0.359 & \cellcolor{sakurapink!15}0.309 & 0.030 & 0.225 & 0.157 \\
     & \cellcolor{royalblue!30}LA & \cellcolor{royalblue!30}0.194 & \cellcolor{royalblue!30}0.403 & \cellcolor{royalblue!30}0.309 & \cellcolor{royalblue!30}0.121 & \cellcolor{royalblue!30}0.293 & \cellcolor{royalblue!30}0.266 & \cellcolor{royalblue!30}0.147 & \cellcolor{royalblue!30}0.281 & \cellcolor{royalblue!30}0.276 & \cellcolor{royalblue!30}0.126 & \cellcolor{royalblue!30}0.267 & \cellcolor{royalblue!30}0.278 & \cellcolor{royalblue!30}0.146 & \cellcolor{royalblue!30}0.367 & \cellcolor{royalblue!30}0.270 & \cellcolor{royalblue!30}0.110 & \cellcolor{royalblue!30}0.312 & \cellcolor{royalblue!30}0.254 \\
     & \cellcolor{myyellow!30}HC & \cellcolor{myyellow!30}0.186 & \cellcolor{myyellow!30}0.389 & \cellcolor{myyellow!30}0.335 & 0.060 & 0.264 & 0.201 & \cellcolor{myyellow!30}0.119 & \cellcolor{myyellow!30}0.226 & \cellcolor{myyellow!30}0.275 & 0.094 & \cellcolor{myyellow!30}0.213 & 0.222 & 0.129 & \cellcolor{myyellow!30}0.330 & \cellcolor{myyellow!30}0.286 & 0.062 & \cellcolor{myyellow!30}0.264 & 0.190 \\
     & \cellcolor{myred!20}LC & 0.105 & 0.304 & 0.234 & 0.028 & 0.232 & 0.163 & 0.093 & \cellcolor{myred!20}0.212 & 0.216 & 0.065 & 0.196 & 0.191 & 0.060 & 0.248 & 0.186 & -0.003 & 0.196 & 0.111 \\
     & \cellcolor{emraldgreen!15}HE & \cellcolor{emraldgreen!15}0.198 & \cellcolor{emraldgreen!15}0.384 & \cellcolor{emraldgreen!15}0.336 & \cellcolor{emraldgreen!15}0.153 & \cellcolor{emraldgreen!15}0.348 & \cellcolor{emraldgreen!15}0.316 & 0.105 & \cellcolor{emraldgreen!15}0.259 & \cellcolor{emraldgreen!15}0.271 & \cellcolor{emraldgreen!15}0.109 & \cellcolor{emraldgreen!15}0.225 & \cellcolor{emraldgreen!15}0.292 & 0.134 & \cellcolor{emraldgreen!15}0.319 & \cellcolor{emraldgreen!15}0.278 & \cellcolor{emraldgreen!15}0.070 & \cellcolor{emraldgreen!15}0.275 & \cellcolor{emraldgreen!15}0.232 \\
     & \cellcolor{milktea!30}LE & 0.150 & \cellcolor{milktea!30}0.355 & 0.267 & 0.084 & \cellcolor{milktea!30}0.290 & 0.231 & \cellcolor{milktea!30}0.132 & \cellcolor{milktea!30}0.230 & 0.253 & \cellcolor{milktea!30}0.099 & \cellcolor{milktea!30}0.236 & 0.238 & 0.111 & \cellcolor{milktea!30}0.305 & 0.223 & \cellcolor{milktea!30}0.082 & \cellcolor{milktea!30}0.288 & \cellcolor{milktea!30}0.218 \\
     & \cellcolor{senzaimidori!20}HN & \cellcolor{senzaimidori!20}0.164 & \cellcolor{senzaimidori!20}0.378 & \cellcolor{senzaimidori!20}0.295 & \cellcolor{senzaimidori!20}0.189 & \cellcolor{senzaimidori!20}0.390 & \cellcolor{senzaimidori!20}0.331 & \cellcolor{senzaimidori!20}0.141 & \cellcolor{senzaimidori!20}0.275 & \cellcolor{senzaimidori!20}0.275 & \cellcolor{senzaimidori!20}0.163 & \cellcolor{senzaimidori!20}0.307 & \cellcolor{senzaimidori!20}0.337 & 0.113 & \cellcolor{senzaimidori!20}0.338 & 0.243 & \cellcolor{senzaimidori!20}0.139 & \cellcolor{senzaimidori!20}0.373 & \cellcolor{senzaimidori!20}0.292 \\
     & \cellcolor{mikan!15}LN & \cellcolor{mikan!15}0.181 & \cellcolor{mikan!15}0.354 & \cellcolor{mikan!15}0.309 & 0.044 & 0.237 & 0.189 & \cellcolor{mikan!15}0.121 & \cellcolor{mikan!15}0.213 & \cellcolor{mikan!15}0.270 & 0.064 & 0.181 & 0.209 & \cellcolor{mikan!15}0.137 & \cellcolor{mikan!15}0.315 & \cellcolor{mikan!15}0.272 & 0.027 & 0.231 & 0.170 \\
     & \cellcolor{myviolet!15}HO & 0.126 & 0.308 & 0.244 & 0.077 & \cellcolor{myviolet!15}0.291 & \cellcolor{myviolet!15}0.249 & 0.098 & 0.174 & 0.240 & 0.084 & \cellcolor{myviolet!15}0.233 & \cellcolor{myviolet!15}0.259 & 0.095 & 0.248 & 0.231 & 0.049 & \cellcolor{myviolet!15}0.261 & \cellcolor{myviolet!15}0.212 \\
     & \cellcolor{mylightblue!15}LO & 0.131 & 0.299 & 0.245 & \cellcolor{mylightblue!15}0.132 & \cellcolor{mylightblue!15}0.340 & \cellcolor{mylightblue!15}0.273 & 0.111 & 0.188 & 0.234 & \cellcolor{mylightblue!15}0.122 & \cellcolor{mylightblue!15}0.236 & \cellcolor{mylightblue!15}0.270 & 0.092 & 0.248 & 0.209 & \cellcolor{mylightblue!15}0.093 & \cellcolor{mylightblue!15}0.295 & \cellcolor{mylightblue!15}0.229 \\
    \hline
    \multirow{11}{*}{L.70b}
     & \cellcolor{gray!15}Df. & & & & \cellcolor{gray!15}0.204 & \cellcolor{gray!15}0.432 & \cellcolor{gray!15}0.373 & & & & \cellcolor{gray!15}0.162 & \cellcolor{gray!15}0.396 & \cellcolor{gray!15}0.330 & & & & \cellcolor{gray!15}0.230 & \cellcolor{gray!15}0.439 & \cellcolor{gray!15}0.386 \\
     & \cellcolor{sakurapink!15}HA & & & & 0.177 & 0.404 & 0.349 & & & & 0.131 & 0.366 & 0.305 & & & & 0.187 & 0.407 & 0.358 \\
     & \cellcolor{royalblue!30}LA & & & & \cellcolor{royalblue!30}0.212 & \cellcolor{royalblue!30}0.444 & \cellcolor{royalblue!30}0.383 & & & & \cellcolor{royalblue!30}0.171 & 0.389 & \cellcolor{royalblue!30}0.337 & & & & \cellcolor{royalblue!30}0.237 & \cellcolor{royalblue!30}0.442 & \cellcolor{royalblue!30}0.387 \\
     & \cellcolor{myyellow!30}HC & & & & 0.185 & 0.414 & 0.354 & & & & 0.147 & 0.382 & 0.317 & & & & 0.202 & 0.422 & 0.366 \\
     & \cellcolor{myred!20}LC & & & & 0.148 & 0.350 & 0.314 & & & & 0.125 & 0.324 & 0.294 & & & & 0.159 & 0.375 & 0.341 \\
     & \cellcolor{emraldgreen!15}HE & & & & 0.184 & 0.380 & 0.332 & & & & 0.122 & 0.361 & 0.293 & & & & 0.185 & 0.406 & 0.348 \\
     & \cellcolor{milktea!30}LE & & & & 0.188 & 0.426 & 0.365 & & & & 0.158 & 0.389 & 0.327 & & & & 0.223 & 0.436 & 0.384 \\
     & \cellcolor{senzaimidori!20}HN & & & & \cellcolor{senzaimidori!20}0.238 & \cellcolor{senzaimidori!20}0.469 & \cellcolor{senzaimidori!20}0.392 & & & & \cellcolor{senzaimidori!20}0.213 & 0.386 & \cellcolor{senzaimidori!20}0.367 & & & & \cellcolor{senzaimidori!20}0.252 & \cellcolor{senzaimidori!20}0.461 & \cellcolor{senzaimidori!20}0.395 \\
     & \cellcolor{mikan!15}LN & & & & 0.184 & 0.405 & 0.353 & & & & 0.137 & 0.382 & 0.309 & & & & 0.198 & 0.417 & 0.361 \\
     & \cellcolor{myviolet!15}HO & & & & 0.148 & 0.366 & 0.319 & & & & 0.098 & 0.349 & 0.277 & & & & 0.166 & 0.402 & 0.340 \\
     & \cellcolor{mylightblue!15}LO & & & & \cellcolor{mylightblue!15}0.213 & \cellcolor{mylightblue!15}0.443 & \cellcolor{mylightblue!15}0.383 & & & & 0.154 & 0.394 & 0.323 & & & & 0.228 & 0.434 & \cellcolor{mylightblue!15}0.387 \\
    \hline
    \multirow{11}{*}{DS}
     & \cellcolor{gray!15}Df. & \cellcolor{gray!15}0.275 & \cellcolor{gray!15}0.478 & \cellcolor{gray!15}0.381 & \cellcolor{gray!15}0.246 & \cellcolor{gray!15}0.506 & \cellcolor{gray!15}0.387 & \cellcolor{gray!15}0.228 & \cellcolor{gray!15}0.416 & \cellcolor{gray!15}0.353 & \cellcolor{gray!15}0.233 & \cellcolor{gray!15}0.441 & \cellcolor{gray!15}0.374 & \cellcolor{gray!15}0.314 & \cellcolor{gray!15}0.468 & \cellcolor{gray!15}0.398 & \cellcolor{gray!15}0.299 & \cellcolor{gray!15}0.510 & \cellcolor{gray!15}0.420 \\
     & \cellcolor{sakurapink!15}HA & 0.264 & \cellcolor{sakurapink!15}0.488 & \cellcolor{sakurapink!15}0.393 & \cellcolor{sakurapink!15}0.249 & 0.503 & 0.387 & \cellcolor{sakurapink!15}0.229 & \cellcolor{sakurapink!15}0.443 & \cellcolor{sakurapink!15}0.361 & 0.210 & 0.431 & 0.358 & \cellcolor{sakurapink!15}0.320 & \cellcolor{sakurapink!15}0.496 & \cellcolor{sakurapink!15}0.411 & 0.291 & 0.505 & 0.414 \\
     & \cellcolor{royalblue!30}LA & \cellcolor{royalblue!30}0.308 & \cellcolor{royalblue!30}0.539 & \cellcolor{royalblue!30}0.419 & \cellcolor{royalblue!30}0.262 & \cellcolor{royalblue!30}0.521 & \cellcolor{royalblue!30}0.402 & \cellcolor{royalblue!30}0.275 & \cellcolor{royalblue!30}0.448 & \cellcolor{royalblue!30}0.397 & \cellcolor{royalblue!30}0.246 & \cellcolor{royalblue!30}0.457 & \cellcolor{royalblue!30}0.383 & \cellcolor{royalblue!30}0.362 & \cellcolor{royalblue!30}0.508 & \cellcolor{royalblue!30}0.448 & \cellcolor{royalblue!30}0.320 & \cellcolor{royalblue!30}0.519 & \cellcolor{royalblue!30}0.431 \\
     & \cellcolor{myyellow!30}HC & \cellcolor{myyellow!30}0.278 & \cellcolor{myyellow!30}0.494 & \cellcolor{myyellow!30}0.390 & \cellcolor{myyellow!30}0.249 & \cellcolor{myyellow!30}0.509 & \cellcolor{myyellow!30}0.394 & \cellcolor{myyellow!30}0.239 & \cellcolor{myyellow!30}0.429 & \cellcolor{myyellow!30}0.364 & \cellcolor{myyellow!30}0.240 & \cellcolor{myyellow!30}0.446 & \cellcolor{myyellow!30}0.379 & \cellcolor{myyellow!30}0.343 & \cellcolor{myyellow!30}0.497 & \cellcolor{myyellow!30}0.422 & \cellcolor{myyellow!30}0.304 & \cellcolor{myyellow!30}0.516 & \cellcolor{myyellow!30}0.429 \\
     & \cellcolor{myred!20}LC & \cellcolor{myred!20}0.277 & \cellcolor{myred!20}0.500 & \cellcolor{myred!20}0.401 & \cellcolor{myred!20}0.249 & \cellcolor{myred!20}0.511 & \cellcolor{myred!20}0.393 & \cellcolor{myred!20}0.242 & \cellcolor{myred!20}0.447 & \cellcolor{myred!20}0.375 & \cellcolor{myred!20}0.238 & \cellcolor{myred!20}0.448 & \cellcolor{myred!20}0.379 & \cellcolor{myred!20}0.337 & \cellcolor{myred!20}0.512 & \cellcolor{myred!20}0.431 & \cellcolor{myred!20}0.306 & \cellcolor{myred!20}0.515 & \cellcolor{myred!20}0.424 \\
     & \cellcolor{emraldgreen!15}HE & \cellcolor{emraldgreen!15}0.279 & 0.473 & 0.375 & \cellcolor{emraldgreen!15}0.260 & \cellcolor{emraldgreen!15}0.514 & \cellcolor{emraldgreen!15}0.397 & 0.221 & \cellcolor{emraldgreen!15}0.421 & 0.334 & \cellcolor{emraldgreen!15}0.243 & \cellcolor{emraldgreen!15}0.447 & \cellcolor{emraldgreen!15}0.380 & \cellcolor{emraldgreen!15}0.333 & \cellcolor{emraldgreen!15}0.481 & 0.398 & \cellcolor{emraldgreen!15}0.314 & 0.509 & \cellcolor{emraldgreen!15}0.429 \\
     & \cellcolor{milktea!30}LE & \cellcolor{milktea!30}0.286 & \cellcolor{milktea!30}0.518 & \cellcolor{milktea!30}0.408 & \cellcolor{milktea!30}0.264 & \cellcolor{milktea!30}0.519 & \cellcolor{milktea!30}0.400 & \cellcolor{milktea!30}0.258 & \cellcolor{milktea!30}0.459 & \cellcolor{milktea!30}0.386 & \cellcolor{milktea!30}0.243 & \cellcolor{milktea!30}0.457 & \cellcolor{milktea!30}0.377 & \cellcolor{milktea!30}0.350 & \cellcolor{milktea!30}0.519 & \cellcolor{milktea!30}0.442 & \cellcolor{milktea!30}0.321 & \cellcolor{milktea!30}0.523 & \cellcolor{milktea!30}0.434 \\
     & \cellcolor{senzaimidori!20}HN & \cellcolor{senzaimidori!20}0.299 & \cellcolor{senzaimidori!20}0.527 & \cellcolor{senzaimidori!20}0.419 & \cellcolor{senzaimidori!20}0.261 & \cellcolor{senzaimidori!20}0.526 & \cellcolor{senzaimidori!20}0.397 & \cellcolor{senzaimidori!20}0.256 & \cellcolor{senzaimidori!20}0.452 & \cellcolor{senzaimidori!20}0.384 & \cellcolor{senzaimidori!20}0.241 & \cellcolor{senzaimidori!20}0.447 & \cellcolor{senzaimidori!20}0.381 & \cellcolor{senzaimidori!20}0.348 & \cellcolor{senzaimidori!20}0.512 & \cellcolor{senzaimidori!20}0.436 & 0.295 & 0.503 & 0.416 \\
     & \cellcolor{mikan!15}LN & \cellcolor{mikan!15}0.285 & \cellcolor{mikan!15}0.503 & \cellcolor{mikan!15}0.403 & \cellcolor{mikan!15}0.252 & \cellcolor{mikan!15}0.511 & \cellcolor{mikan!15}0.397 & \cellcolor{mikan!15}0.245 & \cellcolor{mikan!15}0.447 & \cellcolor{mikan!15}0.377 & 0.232 & 0.441 & \cellcolor{mikan!15}0.375 & \cellcolor{mikan!15}0.347 & \cellcolor{mikan!15}0.507 & \cellcolor{mikan!15}0.436 & \cellcolor{mikan!15}0.309 & \cellcolor{mikan!15}0.521 & \cellcolor{mikan!15}0.430 \\
     & \cellcolor{myviolet!15}HO & 0.271 & \cellcolor{myviolet!15}0.489 & \cellcolor{myviolet!15}0.389 & 0.246 & 0.506 & \cellcolor{myviolet!15}0.395 & 0.221 & \cellcolor{myviolet!15}0.434 & 0.353 & 0.225 & \cellcolor{myviolet!15}0.445 & 0.367 & \cellcolor{myviolet!15}0.327 & \cellcolor{myviolet!15}0.493 & \cellcolor{myviolet!15}0.416 & \cellcolor{myviolet!15}0.305 & 0.506 & \cellcolor{myviolet!15}0.423 \\
     & \cellcolor{mylightblue!15}LO & \cellcolor{mylightblue!15}0.292 & \cellcolor{mylightblue!15}0.525 & \cellcolor{mylightblue!15}0.412 & \cellcolor{mylightblue!15}0.270 & \cellcolor{mylightblue!15}0.528 & \cellcolor{mylightblue!15}0.407 & \cellcolor{mylightblue!15}0.256 & \cellcolor{mylightblue!15}0.448 & \cellcolor{mylightblue!15}0.390 & \cellcolor{mylightblue!15}0.255 & \cellcolor{mylightblue!15}0.462 & \cellcolor{mylightblue!15}0.386 & \cellcolor{mylightblue!15}0.357 & \cellcolor{mylightblue!15}0.509 & \cellcolor{mylightblue!15}0.445 & \cellcolor{mylightblue!15}0.318 & \cellcolor{mylightblue!15}0.524 & \cellcolor{mylightblue!15}0.436 \\
    \hline
    \end{tabular}
\end{table*}
\begin{table*}[htbp]
    \scriptsize
    \centering
    \caption{Overall performance of five LLMs under w/CoT and w/o CoT across datasets (RO, RU, HMR metrics). Cells are shaded gray for the default condition and highlighted with color if performance exceeds the default.}
    \label{tab:conf_res}
    \begin{tabular}{c|c|ccc|ccc|ccc|ccc|ccc|ccc}
    \hline
    \multirow{3}{*}{\textbf{Model}} & \multirow{3}{*}{\textbf{P.}}
    & \multicolumn{6}{c|}{\textbf{LLMJudge}}
    & \multicolumn{6}{c|}{\textbf{TRDL19}}
    & \multicolumn{6}{c}{\textbf{TRDL20}} \\
    \cline{3-20}
     & & \multicolumn{3}{c|}{\textbf{w/CoT}} & \multicolumn{3}{c|}{\textbf{w/o CoT}}
     & \multicolumn{3}{c|}{\textbf{w/CoT}} & \multicolumn{3}{c|}{\textbf{w/o CoT}}
     & \multicolumn{3}{c|}{\textbf{w/CoT}} & \multicolumn{3}{c}{\textbf{w/o CoT}} \\
    \cline{3-20}
     &
     & \textbf{RO} & \textbf{RU} & \textbf{HMR} & \textbf{RO} & \textbf{RU} & \textbf{HMR}
     & \textbf{RO} & \textbf{RU} & \textbf{HMR} & \textbf{RO} & \textbf{RU} & \textbf{HMR}
     & \textbf{RO} & \textbf{RU} & \textbf{HMR} & \textbf{RO} & \textbf{RU} & \textbf{HMR} \\
    \hline
    \multirow{11}{*}{G.}
     & \cellcolor{gray!15}Df. & & & & \cellcolor{gray!15}0.171 & \cellcolor{gray!15}0.932 & \cellcolor{gray!15}0.289 & & & & \cellcolor{gray!15}0.109 & \cellcolor{gray!15}0.960 & \cellcolor{gray!15}0.195 & & & & \cellcolor{gray!15}0.115 & \cellcolor{gray!15}0.974 & \cellcolor{gray!15}0.206 \\
     & \cellcolor{sakurapink!15}HA & & & & 0.136 & \cellcolor{sakurapink!15}0.953 & 0.238 & & & & \cellcolor{sakurapink!15}0.126 & 0.950 & \cellcolor{sakurapink!15}0.223 & & & & \cellcolor{sakurapink!15}0.145 & 0.967 & \cellcolor{sakurapink!15}0.252 \\
     & \cellcolor{royalblue!30}LA & & & & 0.139 & \cellcolor{royalblue!30}0.962 & 0.244 & & & & \cellcolor{royalblue!30}0.122 & 0.958 & \cellcolor{royalblue!30}0.216 & & & & \cellcolor{royalblue!30}0.132 & 0.971 & \cellcolor{royalblue!30}0.232 \\
     & \cellcolor{myyellow!30}HC & & & & 0.095 & \cellcolor{myyellow!30}0.972 & 0.174 & & & & 0.082 & \cellcolor{myyellow!30}0.971 & 0.152 & & & & 0.086 & \cellcolor{myyellow!30}0.981 & 0.158 \\
     & \cellcolor{myred!20}LC & & & & \cellcolor{myred!20}0.178 & \cellcolor{myred!20}0.940 & \cellcolor{myred!20}0.299 & & & & \cellcolor{myred!20}0.163 & 0.936 & \cellcolor{myred!20}0.278 & & & & \cellcolor{myred!20}0.200 & 0.953 & \cellcolor{myred!20}0.330 \\
     & \cellcolor{emraldgreen!15}HE & & & & 0.146 & \cellcolor{emraldgreen!15}0.952 & 0.253 & & & & \cellcolor{emraldgreen!15}0.123 & 0.951 & \cellcolor{emraldgreen!15}0.218 & & & & \cellcolor{emraldgreen!15}0.139 & 0.969 & \cellcolor{emraldgreen!15}0.243 \\
     & \cellcolor{milktea!30}LE & & & & 0.126 & \cellcolor{milktea!30}0.964 & 0.223 & & & & 0.109 & 0.959 & \cellcolor{milktea!30}0.196 & & & & \cellcolor{milktea!30}0.122 & 0.973 & \cellcolor{milktea!30}0.216 \\
     & \cellcolor{senzaimidori!20}HN & & & & \cellcolor{senzaimidori!20}0.198 & \cellcolor{senzaimidori!20}0.936 & \cellcolor{senzaimidori!20}0.327 & & & & \cellcolor{senzaimidori!20}0.203 & 0.918 & \cellcolor{senzaimidori!20}0.333 & & & & \cellcolor{senzaimidori!20}0.222 & 0.945 & \cellcolor{senzaimidori!20}0.360 \\
     & \cellcolor{mikan!15}LN & & & & 0.094 & \cellcolor{mikan!15}0.972 & 0.172 & & & & 0.077 & \cellcolor{mikan!15}0.971 & 0.142 & & & & 0.083 & \cellcolor{mikan!15}0.982 & 0.153 \\
     & \cellcolor{myviolet!15}HO & & & & 0.120 & \cellcolor{myviolet!15}0.960 & 0.213 & & & & 0.105 & 0.958 & 0.190 & & & & \cellcolor{myviolet!15}0.118 & 0.974 & \cellcolor{myviolet!15}0.211 \\
     & \cellcolor{mylightblue!15}LO & & & & 0.099 & \cellcolor{mylightblue!15}0.971 & 0.179 & & & & 0.084 & \cellcolor{mylightblue!15}0.969 & 0.155 & & & & 0.092 & \cellcolor{mylightblue!15}0.981 & 0.168 \\
    \hline
    \multirow{11}{*}{G.m}
     & \cellcolor{gray!15}Df. & \cellcolor{gray!15}0.190 & \cellcolor{gray!15}0.935 & \cellcolor{gray!15}0.315 & \cellcolor{gray!15}0.240 & \cellcolor{gray!15}0.867 & \cellcolor{gray!15}0.376 & \cellcolor{gray!15}0.169 & \cellcolor{gray!15}0.931 & \cellcolor{gray!15}0.286 & \cellcolor{gray!15}0.185 & \cellcolor{gray!15}0.922 & \cellcolor{gray!15}0.308 & \cellcolor{gray!15}0.223 & \cellcolor{gray!15}0.954 & \cellcolor{gray!15}0.361 & \cellcolor{gray!15}0.226 & \cellcolor{gray!15}0.947 & \cellcolor{gray!15}0.365 \\
     & \cellcolor{sakurapink!15}HA & \cellcolor{sakurapink!15}0.229 & 0.901 & \cellcolor{sakurapink!15}0.365 & 0.238 & 0.862 & 0.373 & \cellcolor{sakurapink!15}0.208 & 0.887 & \cellcolor{sakurapink!15}0.337 & \cellcolor{sakurapink!15}0.237 & 0.889 & \cellcolor{sakurapink!15}0.374 & \cellcolor{sakurapink!15}0.254 & 0.927 & \cellcolor{sakurapink!15}0.398 & \cellcolor{sakurapink!15}0.268 & 0.925 & \cellcolor{sakurapink!15}0.415 \\
     & \cellcolor{royalblue!30}LA & \cellcolor{royalblue!30}0.283 & 0.902 & \cellcolor{royalblue!30}0.430 & \cellcolor{royalblue!30}0.257 & \cellcolor{royalblue!30}0.871 & \cellcolor{royalblue!30}0.397 & \cellcolor{royalblue!30}0.253 & 0.895 & \cellcolor{royalblue!30}0.395 & \cellcolor{royalblue!30}0.283 & 0.881 & \cellcolor{royalblue!30}0.428 & \cellcolor{royalblue!30}0.302 & 0.932 & \cellcolor{royalblue!30}0.456 & \cellcolor{royalblue!30}0.336 & 0.922 & \cellcolor{royalblue!30}0.492 \\
     & \cellcolor{myyellow!30}HC & \cellcolor{myyellow!30}0.211 & 0.917 & \cellcolor{myyellow!30}0.344 & 0.207 & \cellcolor{myyellow!30}0.888 & 0.336 & \cellcolor{myyellow!30}0.181 & 0.907 & \cellcolor{myyellow!30}0.301 & \cellcolor{myyellow!30}0.200 & 0.902 & \cellcolor{myyellow!30}0.328 & \cellcolor{myyellow!30}0.228 & 0.931 & \cellcolor{myyellow!30}0.366 & \cellcolor{myyellow!30}0.233 & 0.930 & \cellcolor{myyellow!30}0.372 \\
     & \cellcolor{myred!20}LC & \cellcolor{myred!20}0.270 & 0.893 & \cellcolor{myred!20}0.415 & \cellcolor{myred!20}0.308 & 0.749 & \cellcolor{myred!20}0.437 & \cellcolor{myred!20}0.238 & 0.887 & \cellcolor{myred!20}0.376 & \cellcolor{myred!20}0.277 & 0.878 & \cellcolor{myred!20}0.422 & \cellcolor{myred!20}0.302 & 0.921 & \cellcolor{myred!20}0.455 & \cellcolor{myred!20}0.324 & 0.911 & \cellcolor{myred!20}0.478 \\
     & \cellcolor{emraldgreen!15}HE & \cellcolor{emraldgreen!15}0.302 & 0.835 & \cellcolor{emraldgreen!15}0.444 & 0.239 & \cellcolor{emraldgreen!15}0.868 & 0.375 & \cellcolor{emraldgreen!15}0.259 & 0.826 & \cellcolor{emraldgreen!15}0.395 & \cellcolor{emraldgreen!15}0.297 & 0.827 & \cellcolor{emraldgreen!15}0.437 & \cellcolor{emraldgreen!15}0.332 & 0.850 & \cellcolor{emraldgreen!15}0.478 & \cellcolor{emraldgreen!15}0.334 & 0.857 & \cellcolor{emraldgreen!15}0.481 \\
     & \cellcolor{milktea!30}LE & \cellcolor{milktea!30}0.209 & 0.928 & \cellcolor{milktea!30}0.341 & 0.240 & \cellcolor{milktea!30}0.868 & 0.376 & \cellcolor{milktea!30}0.184 & 0.924 & \cellcolor{milktea!30}0.306 & \cellcolor{milktea!30}0.219 & 0.907 & \cellcolor{milktea!30}0.352 & 0.216 & 0.948 & 0.351 & \cellcolor{milktea!30}0.255 & 0.938 & \cellcolor{milktea!30}0.401 \\
     & \cellcolor{senzaimidori!20}HN & \cellcolor{senzaimidori!20}0.336 & 0.863 & \cellcolor{senzaimidori!20}0.484 & \cellcolor{senzaimidori!20}0.441 & 0.375 & \cellcolor{senzaimidori!20}0.406 & \cellcolor{senzaimidori!20}0.309 & 0.850 & \cellcolor{senzaimidori!20}0.453 & \cellcolor{senzaimidori!20}0.323 & 0.849 & \cellcolor{senzaimidori!20}0.467 & \cellcolor{senzaimidori!20}0.366 & 0.896 & \cellcolor{senzaimidori!20}0.520 & \cellcolor{senzaimidori!20}0.357 & 0.886 & \cellcolor{senzaimidori!20}0.509 \\
     & \cellcolor{mikan!15}LN & \cellcolor{mikan!15}0.195 & 0.919 & \cellcolor{mikan!15}0.322 & 0.208 & \cellcolor{mikan!15}0.877 & 0.337 & \cellcolor{mikan!15}0.177 & 0.909 & \cellcolor{mikan!15}0.297 & \cellcolor{mikan!15}0.204 & 0.902 & \cellcolor{mikan!15}0.332 & 0.212 & 0.936 & 0.345 & \cellcolor{mikan!15}0.238 & 0.934 & \cellcolor{mikan!15}0.379 \\
     & \cellcolor{myviolet!15}HO & \cellcolor{myviolet!15}0.228 & 0.905 & \cellcolor{myviolet!15}0.364 & 0.234 & \cellcolor{myviolet!15}0.872 & 0.369 & \cellcolor{myviolet!15}0.197 & 0.899 & \cellcolor{myviolet!15}0.324 & \cellcolor{myviolet!15}0.225 & 0.896 & \cellcolor{myviolet!15}0.360 & \cellcolor{myviolet!15}0.251 & 0.930 & \cellcolor{myviolet!15}0.395 & \cellcolor{myviolet!15}0.260 & 0.925 & \cellcolor{myviolet!15}0.406 \\
     & \cellcolor{mylightblue!15}LO & 0.180 & \cellcolor{mylightblue!15}0.936 & 0.302 & 0.220 & \cellcolor{mylightblue!15}0.875 & 0.352 & 0.155 & \cellcolor{mylightblue!15}0.936 & 0.266 & \cellcolor{mylightblue!15}0.188 & 0.920 & \cellcolor{mylightblue!15}0.313 & 0.189 & \cellcolor{mylightblue!15}0.960 & 0.315 & \cellcolor{mylightblue!15}0.235 & 0.945 & \cellcolor{mylightblue!15}0.376 \\
    \hline
    \multirow{11}{*}{L.8b}
     & \cellcolor{gray!15}Df. & \cellcolor{gray!15}0.330 & \cellcolor{gray!15}0.303 & \cellcolor{gray!15}0.316 & \cellcolor{gray!15}0.427 & \cellcolor{gray!15}0.355 & \cellcolor{gray!15}0.388 & \cellcolor{gray!15}0.334 & \cellcolor{gray!15}0.452 & \cellcolor{gray!15}0.384 & \cellcolor{gray!15}0.418 & \cellcolor{gray!15}0.515 & \cellcolor{gray!15}0.461 & \cellcolor{gray!15}0.355 & \cellcolor{gray!15}0.282 & \cellcolor{gray!15}0.315 & \cellcolor{gray!15}0.458 & \cellcolor{gray!15}0.302 & \cellcolor{gray!15}0.364 \\
     & \cellcolor{sakurapink!15}HA & \cellcolor{sakurapink!15}0.361 & \cellcolor{sakurapink!15}0.307 & \cellcolor{sakurapink!15}0.332 & \cellcolor{sakurapink!15}0.450 & \cellcolor{sakurapink!15}0.436 & \cellcolor{sakurapink!15}0.443 & \cellcolor{sakurapink!15}0.365 & 0.433 & \cellcolor{sakurapink!15}0.396 & \cellcolor{sakurapink!15}0.434 & \cellcolor{sakurapink!15}0.576 & \cellcolor{sakurapink!15}0.495 & \cellcolor{sakurapink!15}0.373 & 0.275 & \cellcolor{sakurapink!15}0.316 & \cellcolor{sakurapink!15}0.474 & \cellcolor{sakurapink!15}0.352 & \cellcolor{sakurapink!15}0.404 \\
     & \cellcolor{royalblue!30}LA & \cellcolor{royalblue!30}0.564 & 0.262 & \cellcolor{royalblue!30}0.358 & \cellcolor{royalblue!30}0.620 & 0.281 & 0.387 & \cellcolor{royalblue!30}0.569 & 0.386 & \cellcolor{royalblue!30}0.460 & \cellcolor{royalblue!30}0.606 & 0.404 & \cellcolor{royalblue!30}0.484 & \cellcolor{royalblue!30}0.571 & 0.254 & \cellcolor{royalblue!30}0.352 & \cellcolor{royalblue!30}0.632 & 0.289 & \cellcolor{royalblue!30}0.396 \\
     & \cellcolor{myyellow!30}HC & 0.278 & \cellcolor{myyellow!30}0.345 & 0.308 & 0.410 & \cellcolor{myyellow!30}0.424 & \cellcolor{myyellow!30}0.417 & 0.295 & \cellcolor{myyellow!30}0.497 & 0.370 & 0.344 & \cellcolor{myyellow!30}0.585 & 0.433 & 0.306 & \cellcolor{myyellow!30}0.299 & 0.302 & 0.402 & \cellcolor{myyellow!30}0.350 & \cellcolor{myyellow!30}0.374 \\
     & \cellcolor{myred!20}LC & \cellcolor{myred!20}0.528 & \cellcolor{myred!20}0.340 & \cellcolor{myred!20}0.414 & \cellcolor{myred!20}0.542 & \cellcolor{myred!20}0.421 & \cellcolor{myred!20}0.474 & \cellcolor{myred!20}0.533 & \cellcolor{myred!20}0.453 & \cellcolor{myred!20}0.490 & \cellcolor{myred!20}0.556 & 0.502 & \cellcolor{myred!20}0.528 & \cellcolor{myred!20}0.546 & \cellcolor{myred!20}0.320 & \cellcolor{myred!20}0.403 & \cellcolor{myred!20}0.578 & \cellcolor{myred!20}0.431 & \cellcolor{myred!20}0.494 \\
     & \cellcolor{emraldgreen!15}HE & 0.263 & \cellcolor{emraldgreen!15}0.319 & 0.288 & 0.276 & 0.355 & 0.310 & 0.268 & \cellcolor{emraldgreen!15}0.459 & 0.338 & 0.292 & 0.489 & 0.366 & 0.289 & 0.282 & 0.285 & 0.376 & \cellcolor{emraldgreen!15}0.345 & 0.360 \\
     & \cellcolor{milktea!30}LE & \cellcolor{milktea!30}0.404 & \cellcolor{milktea!30}0.323 & \cellcolor{milktea!30}0.359 & \cellcolor{milktea!30}0.520 & 0.341 & \cellcolor{milktea!30}0.412 & \cellcolor{milktea!30}0.411 & \cellcolor{milktea!30}0.465 & \cellcolor{milktea!30}0.436 & \cellcolor{milktea!30}0.437 & \cellcolor{milktea!30}0.519 & \cellcolor{milktea!30}0.474 & \cellcolor{milktea!30}0.427 & \cellcolor{milktea!30}0.284 & \cellcolor{milktea!30}0.341 & \cellcolor{milktea!30}0.487 & \cellcolor{milktea!30}0.323 & \cellcolor{milktea!30}0.388 \\
     & \cellcolor{senzaimidori!20}HN & \cellcolor{senzaimidori!20}0.710 & 0.232 & \cellcolor{senzaimidori!20}0.349 & \cellcolor{senzaimidori!20}0.626 & 0.243 & 0.351 & \cellcolor{senzaimidori!20}0.711 & 0.306 & \cellcolor{senzaimidori!20}0.428 & \cellcolor{senzaimidori!20}0.697 & 0.291 & 0.411 & \cellcolor{senzaimidori!20}0.711 & 0.237 & \cellcolor{senzaimidori!20}0.356 & \cellcolor{senzaimidori!20}0.665 & 0.239 & 0.352 \\
     & \cellcolor{mikan!15}LN & 0.290 & \cellcolor{mikan!15}0.307 & 0.298 & 0.423 & \cellcolor{mikan!15}0.435 & \cellcolor{mikan!15}0.429 & 0.308 & \cellcolor{mikan!15}0.460 & 0.369 & 0.378 & \cellcolor{mikan!15}0.602 & \cellcolor{mikan!15}0.464 & 0.302 & 0.256 & 0.277 & 0.440 & \cellcolor{mikan!15}0.382 & \cellcolor{mikan!15}0.409 \\
     & \cellcolor{myviolet!15}HO & \cellcolor{myviolet!15}0.376 & 0.292 & \cellcolor{myviolet!15}0.329 & 0.417 & \cellcolor{myviolet!15}0.368 & \cellcolor{myviolet!15}0.391 & \cellcolor{myviolet!15}0.368 & \cellcolor{myviolet!15}0.472 & \cellcolor{myviolet!15}0.414 & \cellcolor{myviolet!15}0.419 & 0.504 & 0.458 & \cellcolor{myviolet!15}0.405 & 0.253 & 0.312 & \cellcolor{myviolet!15}0.488 & 0.299 & \cellcolor{myviolet!15}0.371 \\
     & \cellcolor{mylightblue!15}LO & \cellcolor{mylightblue!15}0.374 & \cellcolor{mylightblue!15}0.307 & \cellcolor{mylightblue!15}0.337 & \cellcolor{mylightblue!15}0.463 & 0.313 & 0.373 & \cellcolor{mylightblue!15}0.380 & \cellcolor{mylightblue!15}0.472 & \cellcolor{mylightblue!15}0.421 & \cellcolor{mylightblue!15}0.421 & 0.477 & 0.447 & \cellcolor{mylightblue!15}0.397 & 0.263 & \cellcolor{mylightblue!15}0.316 & 0.457 & 0.291 & 0.355 \\
    \hline
    \multirow{11}{*}{L.70b}
     & \cellcolor{gray!15}Df. & & & & \cellcolor{gray!15}0.210 & \cellcolor{gray!15}0.932 & \cellcolor{gray!15}0.343 & & & & \cellcolor{gray!15}0.198 & \cellcolor{gray!15}0.922 & \cellcolor{gray!15}0.326 & & & & \cellcolor{gray!15}0.266 & \cellcolor{gray!15}0.946 & \cellcolor{gray!15}0.415 \\
     & \cellcolor{sakurapink!15}HA & & & & 0.206 & 0.928 & 0.337 & & & & 0.183 & \cellcolor{sakurapink!15}0.923 & 0.305 & & & & 0.238 & \cellcolor{sakurapink!15}0.950 & 0.380 \\
     & \cellcolor{royalblue!30}LA & & & & \cellcolor{royalblue!30}0.238 & 0.920 & \cellcolor{royalblue!30}0.378 & & & & \cellcolor{royalblue!30}0.212 & 0.919 & \cellcolor{royalblue!30}0.345 & & & & \cellcolor{royalblue!30}0.284 & 0.945 & \cellcolor{royalblue!30}0.437 \\
     & \cellcolor{myyellow!30}HC & & & & 0.160 & \cellcolor{myyellow!30}0.951 & 0.273 & & & & 0.147 & \cellcolor{myyellow!30}0.946 & 0.255 & & & & 0.200 & \cellcolor{myyellow!30}0.967 & 0.331 \\
     & \cellcolor{myred!20}LC & & & & \cellcolor{myred!20}0.292 & 0.894 & \cellcolor{myred!20}0.440 & & & & \cellcolor{myred!20}0.273 & 0.878 & \cellcolor{myred!20}0.416 & & & & \cellcolor{myred!20}0.320 & 0.925 & \cellcolor{myred!20}0.475 \\
     & \cellcolor{emraldgreen!15}HE & & & & 0.176 & \cellcolor{emraldgreen!15}0.954 & 0.298 & & & & 0.155 & \cellcolor{emraldgreen!15}0.946 & 0.267 & & & & 0.220 & \cellcolor{emraldgreen!15}0.968 & 0.358 \\
     & \cellcolor{milktea!30}LE & & & & 0.202 & 0.924 & 0.331 & & & & 0.182 & 0.917 & 0.304 & & & & 0.232 & 0.946 & 0.373 \\
     & \cellcolor{senzaimidori!20}HN & & & & \cellcolor{senzaimidori!20}0.317 & 0.843 & \cellcolor{senzaimidori!20}0.460 & & & & \cellcolor{senzaimidori!20}0.293 & 0.859 & \cellcolor{senzaimidori!20}0.437 & & & & \cellcolor{senzaimidori!20}0.357 & 0.873 & \cellcolor{senzaimidori!20}0.507 \\
     & \cellcolor{mikan!15}LN & & & & 0.167 & \cellcolor{mikan!15}0.946 & 0.284 & & & & 0.152 & \cellcolor{mikan!15}0.940 & 0.262 & & & & 0.181 & \cellcolor{mikan!15}0.968 & 0.305 \\
     & \cellcolor{myviolet!15}HO & & & & \cellcolor{myviolet!15}0.232 & 0.925 & \cellcolor{myviolet!15}0.371 & & & & \cellcolor{myviolet!15}0.204 & 0.913 & \cellcolor{myviolet!15}0.334 & & & & \cellcolor{myviolet!15}0.275 & 0.946 & \cellcolor{myviolet!15}0.426 \\
     & \cellcolor{mylightblue!15}LO & & & & \cellcolor{mylightblue!15}0.230 & 0.924 & \cellcolor{mylightblue!15}0.369 & & & & \cellcolor{mylightblue!15}0.216 & 0.919 & \cellcolor{mylightblue!15}0.349 & & & & \cellcolor{mylightblue!15}0.291 & \cellcolor{mylightblue!15}0.947 & \cellcolor{mylightblue!15}0.445 \\
    \hline
    \multirow{11}{*}{DS}
     & \cellcolor{gray!15}Df. & \cellcolor{gray!15}0.077 & \cellcolor{gray!15}0.982 & \cellcolor{gray!15}0.142 & \cellcolor{gray!15}0.142 & \cellcolor{gray!15}0.939 & \cellcolor{gray!15}0.247 & \cellcolor{gray!15}0.096 & \cellcolor{gray!15}0.963 & \cellcolor{gray!15}0.175 & \cellcolor{gray!15}0.142 & \cellcolor{gray!15}0.923 & \cellcolor{gray!15}0.246 & \cellcolor{gray!15}0.112 & \cellcolor{gray!15}0.978 & \cellcolor{gray!15}0.201 & \cellcolor{gray!15}0.136 & \cellcolor{gray!15}0.960 & \cellcolor{gray!15}0.238 \\
     & \cellcolor{sakurapink!15}HA & \cellcolor{sakurapink!15}0.093 & 0.972 & \cellcolor{sakurapink!15}0.169 & 0.130 & \cellcolor{sakurapink!15}0.945 & 0.229 & \cellcolor{sakurapink!15}0.104 & 0.952 & \cellcolor{sakurapink!15}0.188 & 0.119 & \cellcolor{sakurapink!15}0.935 & 0.211 & \cellcolor{sakurapink!15}0.114 & 0.968 & \cellcolor{sakurapink!15}0.204 & 0.125 & \cellcolor{sakurapink!15}0.965 & 0.221 \\
     & \cellcolor{royalblue!30}LA & 0.077 & 0.977 & \cellcolor{royalblue!30}0.143 & 0.125 & \cellcolor{royalblue!30}0.943 & 0.220 & 0.093 & 0.954 & 0.169 & 0.116 & \cellcolor{royalblue!30}0.939 & 0.206 & 0.101 & 0.970 & 0.184 & 0.118 & \cellcolor{royalblue!30}0.968 & 0.210 \\
     & \cellcolor{myyellow!30}HC & 0.074 & 0.980 & 0.137 & 0.120 & \cellcolor{myyellow!30}0.952 & 0.213 & 0.094 & 0.957 & 0.171 & 0.114 & \cellcolor{myyellow!30}0.942 & 0.204 & 0.105 & 0.971 & 0.190 & 0.119 & \cellcolor{myyellow!30}0.971 & 0.212 \\
     & \cellcolor{myred!20}LC & \cellcolor{myred!20}0.100 & 0.970 & \cellcolor{myred!20}0.181 & \cellcolor{myred!20}0.158 & 0.931 & \cellcolor{myred!20}0.271 & \cellcolor{myred!20}0.116 & 0.940 & \cellcolor{myred!20}0.206 & \cellcolor{myred!20}0.159 & 0.913 & \cellcolor{myred!20}0.270 & \cellcolor{myred!20}0.128 & 0.956 & \cellcolor{myred!20}0.226 & \cellcolor{myred!20}0.154 & 0.955 & \cellcolor{myred!20}0.266 \\
     & \cellcolor{emraldgreen!15}HE & 0.070 & 0.980 & 0.131 & 0.127 & \cellcolor{emraldgreen!15}0.948 & 0.224 & \cellcolor{emraldgreen!15}0.100 & \cellcolor{emraldgreen!15}0.965 & \cellcolor{emraldgreen!15}0.182 & 0.118 & \cellcolor{emraldgreen!15}0.938 & 0.209 & 0.106 & \cellcolor{emraldgreen!15}0.980 & 0.192 & 0.117 & \cellcolor{emraldgreen!15}0.968 & 0.209 \\
     & \cellcolor{milktea!30}LE & \cellcolor{milktea!30}0.094 & 0.973 & \cellcolor{milktea!30}0.171 & 0.140 & \cellcolor{milktea!30}0.940 & 0.244 & 0.096 & 0.957 & 0.175 & 0.128 & \cellcolor{milktea!30}0.929 & 0.225 & 0.098 & 0.974 & 0.178 & 0.130 & \cellcolor{milktea!30}0.962 & 0.230 \\
     & \cellcolor{senzaimidori!20}HN & \cellcolor{senzaimidori!20}0.108 & 0.961 & \cellcolor{senzaimidori!20}0.194 & 0.139 & \cellcolor{senzaimidori!20}0.952 & 0.243 & \cellcolor{senzaimidori!20}0.106 & 0.947 & \cellcolor{senzaimidori!20}0.191 & 0.135 & \cellcolor{senzaimidori!20}0.928 & 0.235 & \cellcolor{senzaimidori!20}0.120 & 0.964 & \cellcolor{senzaimidori!20}0.214 & 0.131 & \cellcolor{senzaimidori!20}0.963 & 0.230 \\
     & \cellcolor{mikan!15}LN & \cellcolor{mikan!15}0.084 & 0.976 & \cellcolor{mikan!15}0.155 & \cellcolor{mikan!15}0.145 & 0.938 & \cellcolor{mikan!15}0.251 & \cellcolor{mikan!15}0.101 & 0.952 & \cellcolor{mikan!15}0.183 & 0.132 & \cellcolor{mikan!15}0.928 & 0.231 & 0.111 & 0.966 & 0.200 & 0.134 & \cellcolor{mikan!15}0.963 & 0.235 \\
     & \cellcolor{myviolet!15}HO & \cellcolor{myviolet!15}0.086 & 0.974 & \cellcolor{myviolet!15}0.159 & 0.134 & \cellcolor{myviolet!15}0.945 & 0.234 & \cellcolor{myviolet!15}0.102 & 0.954 & \cellcolor{myviolet!15}0.185 & 0.135 & \cellcolor{myviolet!15}0.932 & 0.236 & 0.107 & 0.971 & 0.193 & 0.124 & \cellcolor{myviolet!15}0.965 & 0.220 \\
     & \cellcolor{mylightblue!15}LO & \cellcolor{mylightblue!15}0.093 & 0.973 & \cellcolor{mylightblue!15}0.170 & 0.140 & \cellcolor{mylightblue!15}0.940 & 0.243 & \cellcolor{mylightblue!15}0.099 & 0.949 & \cellcolor{mylightblue!15}0.179 & 0.142 & \cellcolor{mylightblue!15}0.928 & 0.246 & 0.109 & 0.967 & 0.196 & 0.129 & \cellcolor{mylightblue!15}0.961 & 0.227 \\
    \hline
    \end{tabular}
\end{table*}
Table~\ref{tab:alignment_res} and Table~\ref{tab:conf_res}s presents the full results of Experiment 1, where  G. denotes GPT-4o, G.m denotes GPT-4o-mini, L.8b denotes Llama-3-8B, L.70b denotes Llama-3-70B, DS denotes DeepSeek-v3, P. denotes personality, and Df. denotes the default setting, i.e., the condition without any personality prompt. 

\section{An Example of the Computation of RO, RU and HMR}
\label{app:hmr}
\begin{table}[ht]
\centering
\caption{A toy example for computing RO, RU, and HMR in relevance assessment}
\label{tab:toy}
\begin{tabular}{cccccc}
\toprule
\textbf{docid} & \textbf{ground-truth Label} & \textbf{Predicted} & \textbf{Confidence} & \textbf{Correct?} \\
\midrule
a & 3 & 3 & 0.90 & \cellcolor{green!15}{\checkmark} \\
b & 2 & 2 & 0.60 &\cellcolor{green!15}{\checkmark} \\
c & 1 & 2 & 0.85 &   \cellcolor{red!15}{$\times$} \\
d & 0 & 0 & 0.40 & \cellcolor{green!15}{\checkmark}\\
e & 2 & 0 & 0.30 &  \cellcolor{red!15}{$\times$}  \\
\bottomrule
\end{tabular}
\end{table}
Table~\ref{tab:toy} presents a simulated relevance assessment scenario in which an assessor evaluates the relevance levels of a set of documents $\{a, b, c, d, e\}$, providing both predicted relevance scores and corresponding (normalized) confidence values for each judgment. We compute the metrics as follows. Let $I^+$ denote the set of correctly predicted instances and $I^-$ the set of incorrect ones.
In this toy example, $I^+ = \{a, b, d\}$ and $I^- = \{c, e\}$.

The overconfidence penalty $O$ and the underconfidence penalty $U$ are defined as:
\begin{align*}
O &= \sum_{i \in I^-} p(i) = 0.85 + 0.30 = 1.15, \\
U &= \sum_{i \in I^+} (1 - p(i)) = (1-0.9)+(1-0.6)+(1-0.4)=1.1.
\end{align*}

The rewards for suppressing overconfidence and underconfidence are:
\begin{align*}
R_O &= 1 - \frac{O}{|I^-|} = 1 - \frac{1.15}{2} = 0.425, \\
R_U &= 1 - \frac{U}{|I^+|} = 1 - \frac{1.1}{3} = 0.633.
\end{align*}

Finally, the Harmonic Mean of Rewards (HMR) is:
\begin{align*}
\text{HMR} &= \frac{2 R_O R_U}{R_O + R_U} = \frac{2 \times 0.425 \times 0.633}{0.425 + 0.633} \approx 0.508.
\end{align*}

Although the model correctly labels three out of five instances, its confidence is miscalibrated:
it is overconfident on wrong predictions (e.g., confidence = 0.85 for ID 3) and underconfident on correct ones (e.g., confidence = 0.40 for ID 4).
The resulting HMR reflects a balanced penalty across both error types, offering a more informative reliability measure than accuracy alone.

\end{document}